%% file: main.tex
\documentclass[lettersize,journal]{IEEEtran}
\usepackage{amsmath,amsfonts}
\usepackage{algorithmic}
\usepackage{algorithm}
\usepackage{array}
\usepackage[caption=false,font=normalsize,labelfont=sf,textfont=sf]{subfig}
\usepackage{textcomp}
\usepackage{stfloats}
\usepackage{url}
\usepackage{verbatim}
\usepackage{graphicx}
\usepackage{booktabs}
\usepackage{cite}
\usepackage{multirow}
\usepackage{placeins}
\usepackage{xcolor}
\usepackage{tabularx}
\usepackage{makecell}
\usepackage{float}

\providecommand{\rev}[1]{#1}
\makeatletter
\def\@IEEEBIOskipN{1\baselineskip}
\makeatother

\begin{document}

\title{DuSRFlow: Dual-Lens Super-Resolution via Parallax-Aware Flow Estimation and Complementary Reference Warping}

\author{Fangpu~Zhang, Huanjing~Yue,~\IEEEmembership{Senior Member,~IEEE}, Zifan~Cui, Shaoxiong~Tu, \\Qirui~Yang, and~Jingyu~Yang,~\IEEEmembership{Senior Member,~IEEE}

\thanks{This work was supported in part by the National Natural Science Foundation of China under Grant 62472308 and Grant 62231018.
F. Zhang and H. Yue (corresponding author) are with the State Key Laboratory of Smart Power Distribution Equipment and System, Tianjin University and School of Electrical and Information Engineering, Tianjin University. Z. Cui, Q. Yang, and J. Yang are with the School of Electrical and Information Engineering, Tianjin University, Tianjin, China, 300072. S. Tu is with Huawei Technologies Co., Ltd.}

\thanks{This paper has supplementary downloadable material available at http://ieeexplore.ieee.org, provided by the authors. The material includes additional implementation details, quantitative and qualitative experimental results, ablation studies, model complexity analysis, and memory overhead evaluation. This material is 1.02 MB in size.
}

}

\markboth{Submitted to IEEE Transactions on Multimedia}
{Shell \MakeLowercase{\textit{et al.}}: A Sample Article Using IEEEtran.cls for IEEE Journals}

\maketitle

\input{0abstract}

\input{1Introduction}

\input{2Related_Works}
\input{3Methods}
\input{4Experiment}

\FloatBarrier
\bibliographystyle{IEEEtran}
\bibliography{Refs}

\clearpage
\input{supplementary}

\end{document}

%% file: 0abstract.tex
\begin{abstract}
Dual-lens super-resolution (DuSR) aims to enhance a wide-angle low-resolution (LR) image by exploiting a telephoto reference (Ref) image. A central challenge in DuSR is reliable reference utilization under large parallax, occlusion, and cross-camera imaging discrepancies. Existing methods often employ off-the-shelf monocular optical flow models to provide geometric priors. However, their assumptions of photometric consistency and smooth motion are violated in dual-lens scenarios, leading to unreliable alignment and degraded reference-based reconstruction. To address these issues, we propose DuSRFlow, a dual-lens super-resolution framework that improves reference utilization from both correspondence learning and reference aggregation. First, we introduce DuFlowNet, a dedicated dual-lens flow estimator for robust cross-view alignment. To train DuFlowNet without ground-truth flow annotations, we develop a real--synthetic data fusion strategy that composites synthetic foreground objects onto real dual-lens image pairs, generating controllable large-parallax samples while preserving realistic cross-camera imaging characteristics. Second, we observe that reference reliability is highly spatially uneven in DuSR, where unreliable local reference responses may suppress the utilization of reliable cues. We therefore propose a Complementary Reference Warping strategy that coordinates patch-based warping and kernel-free-guided deformable warping according to local reference reliability, producing a more reliable reference representation before LR--Ref fusion. In addition, we construct DuSR-RealV2, a challenging real-world benchmark with complex depth variations, large parallax, and occlusion. Extensive experiments on multiple DuSR benchmarks demonstrate that DuSRFlow achieves state-of-the-art performance in reconstruction quality, perceptual quality, and cross-dataset generalization.

\end{abstract}


%% file: 1Introduction.tex
\section{Introduction}
\IEEEPARstart{S}{ingle}-image super-resolution (SISR)~\cite{srcnn2015,rcan2018,swinir2021} aims to reconstruct a high-resolution (HR) image from its low-resolution (LR) observation, but is inherently ill-posed due to irreversible information loss. Reference-based super-resolution (RefSR)~\cite{srntt2019,TTSR2020,crossnet2018,SSEN2020,c2-matching2021,huang2022task,datsr2022,masa2021} alleviates this issue by leveraging an additional reference (Ref) image with correlated textures. In particular, the dual-lens RefSR (DuSR) setting~\cite{DCSR2021,SelfDZSR2022,ZeDuSR2023,geometry2023,kim2023efficient,lee2022reference} naturally provides such references via modern multi-camera systems, where a telephoto image offers high-resolution details for reconstructing a wide-angle LR input. However, exploiting the telephoto reference remains challenging due to dual-lens parallax and cross-camera imaging discrepancies, including differences in resolution, color response, and brightness.

\begin{figure*}[t]
  \centering
  \includegraphics[width=0.9\linewidth]{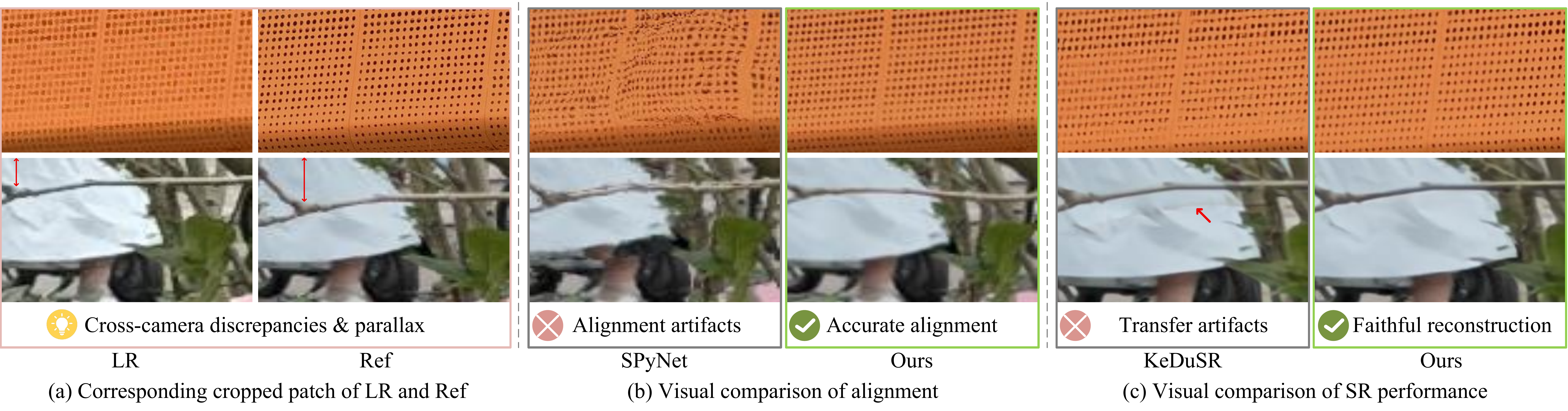}
  \caption{(a) LR--Ref patches suffer from cross-camera  discrepancies and parallax, challenging generic optical flow estimation. (b) Benefiting from real--synthetic data fusion, our dual-lens flow achieves cleaner alignment than SPyNet with fewer warping artifacts. (c) More accurate alignment improves reference utilization, leading to more faithful structures and fewer reference-imprinted artifacts than KeDuSR.}
  \label{fig:motivation}
\end{figure*}

\begin{figure}
  \centering
  \includegraphics[width=0.9\linewidth]{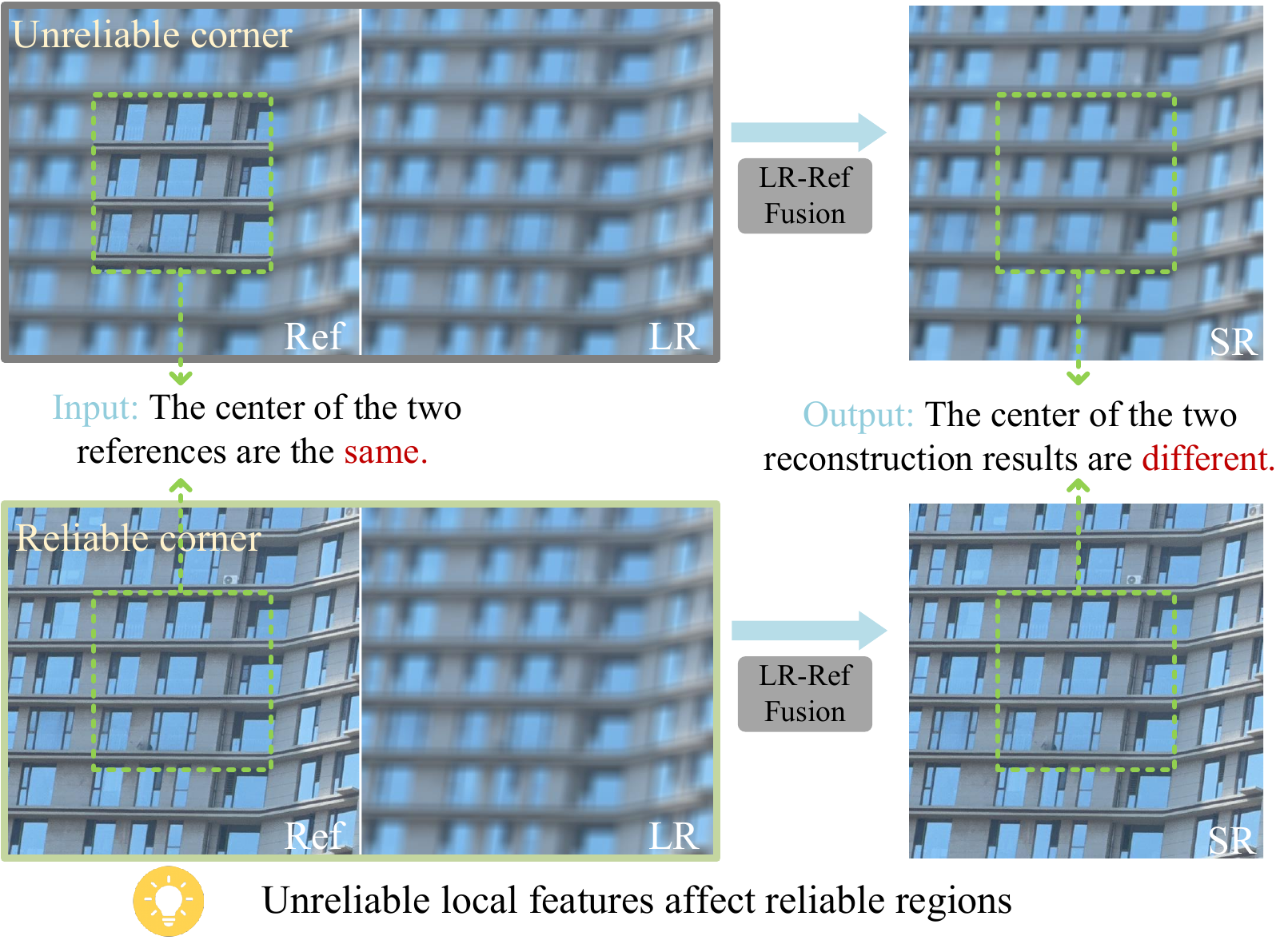}
    \caption{Although the central reference regions have the same quality in both cases,
    the final SR results differ because unreliable corner references reduce the overall reliability of the fused reference representation.}
   \label{fig:reference-reliability}
\end{figure}

To exploit the telephoto reference, DuSR methods need to compensate for dual-lens parallax by warping the Ref image toward the LR view. Existing DuSR methods~\cite{2024KeDuSR} typically employ optical flow models, such as SPyNet~\cite{spynet}, to provide geometric priors before reference matching and fusion. However, this parallax compensation problem differs substantially from conventional monocular optical flow estimation. The LR and Ref images often exhibit depth-dependent displacement and occlusion, while cross-camera discrepancies introduce resolution and appearance gaps that weaken photometric consistency. These factors violate the assumptions of off-the-shelf flow estimators, which are mainly designed for image pairs with similar imaging characteristics and relatively smooth motion patterns. Consequently, generic optical flow models tend to produce unreliable alignments in dual-lens scenarios, degrading subsequent reference utilization and reconstruction quality, as shown in Fig.~\ref{fig:motivation}. Although the flow network can be fine-tuned with the final SR reconstruction loss, such supervision is indirect and acts only on the reconstruction output rather than the intermediate alignment. Therefore, it remains insufficient for learning accurate dual-lens correspondence, especially under large parallax.

This raises a key question: how can we obtain effective supervision for dual-lens correspondence learning under large parallax? Existing paired DuSR datasets are built by registering the telephoto view to the wide-angle view with SIFT-based coarse alignment and optical-flow refinement, which favors samples with small parallax. Large-parallax samples, in contrast, are difficult to align accurately and are therefore insufficiently represented in released paired data. As a result, such data are inadequate for learning robust correspondence in challenging dual-lens scenarios. To address this issue, we introduce \textbf{DuFlowNet}, a dedicated dual-lens flow estimator trained with intermediate alignment supervision. Instead of requiring ground-truth optical flow annotations, we develop a \textbf{real–synthetic data fusion} strategy that provides image-level supervision by constructing view-consistent training samples with controllable parallax. Specifically, synthetic foreground objects are composited onto real wide-angle and telephoto image pairs, introducing controllable large parallax and view-dependent occlusion while preserving realistic background structures and cross-camera discrepancies. In this way, the generated data enrich the geometric variations absent from existing paired datasets, enabling DuFlowNet to learn more reliable correspondence under both severe parallax and cross-camera discrepancies.

Although accurate correspondence provides a stronger geometric basis for reference transfer, it does not guarantee that the transferred reference features are uniformly reliable across the image. Due to the limited overlap between the telephoto reference and the wide-angle LR image, reference reliability is highly spatially uneven: some regions contain highly similar and informative reference cues, whereas others suffer from weak correspondence, occlusion, or missing reference content. This imbalance makes subsequent LR--Ref fusion difficult. Existing methods~\cite{2024KeDuSR,TTSR2020,huang2022task} usually mitigate this issue using adaptive weighting masks and standard convolutional fusion. However, when reliable and unreliable reference responses are mixed within a single Ref branch, the fusion module tends to make conservative use of the reference as a whole. As a result, reliable cues in well-matched regions may still be underutilized, because unreliable local responses in weakly matched regions can affect the overall reference representation, as illustrated in Fig.~\ref{fig:reference-reliability}.

A straightforward remedy is to make LR--Ref fusion more adaptive. However, adapting only the final LR--Ref fusion may still be insufficient, because the reference representation has already been formed by a single warping strategy before being fused with LR features. In practice, different warping mechanisms exhibit complementary advantages: patch-based warping explicitly retrieves highly similar reference patches and preserves faithful structures in well-overlapped regions, but becomes unreliable in low-similarity areas; DCN-based warping provides flexible local compensation in weakly correlated or partially occluded regions, but may compromise fine structures in highly reliable regions. To better exploit these complementary behaviors, we propose a \textbf{Complementary Reference Warping} strategy that improves the reference representation before LR--Ref fusion. Specifically, we first fuse the warped reference features from patch-based warping and DCN-based warping through a confidence-aware mechanism, producing a more balanced and reliable reference representation, and then fuse it with LR features for reconstruction. Instead of merely balancing high- and low-similarity regions, the proposed design reduces the adverse influence of unreliable reference responses and better exploits the complementary strengths of both warping branches, leading to more effective use of reliable reference cues across regions with different similarity levels.

These challenges also expose the limitations of existing DuSR benchmarks. 
Since reliable correspondence estimation and reference aggregation are most difficult under large parallax, occlusion, and complex depth variations, a practical benchmark should sufficiently cover such challenging cases. 
However, existing datasets either rely on synthetic degradations or are affected by misalignment, motion artifacts, and limited depth diversity, making them insufficient for comprehensive evaluation in real dual-lens scenarios. 
To bridge this gap, we construct \textbf{DuSR-RealV2}, a new real-world benchmark with complex depth variations and large parallax. 
It contains well-aligned paired data for quantitative evaluation and challenging unpaired data for assessing generalization in more realistic settings.

In summary, the contributions of this work are as follows:

\begin{itemize}
    \item We propose \textbf{DuSRFlow}, a dual-lens super-resolution framework that improves reference utilization under large parallax and cross-camera discrepancies.

    \item We introduce a \textbf{real--synthetic data fusion} strategy to generate controllable large-parallax samples, enabling the training of DuFlowNet, a dedicated dual-lens flow estimator for robust cross-view alignment under large parallax and cross-camera discrepancies, without ground-truth flow.
    
    \item We propose a \textbf{Complementary Reference Warping} strategy that coordinates patch-based and DCN-based warping according to local reference reliability, alleviating unreliable reference transfer and enhancing the utilization of reliable reference cues.

    \item We construct \textbf{DuSR-RealV2}, a challenging real-world benchmark with complex depth variations, large parallax, and occlusion, enabling a more comprehensive evaluation of real-world dual-lens super-resolution.

\end{itemize}

%% file: 2Related_Works.tex
\section{Related Work}

\subsection{Reference-Based Super-Resolution}
Compared with SISR~\cite{rcan2018,swinir2021,esrgan2018,bsrgan2021}, RefSR exploits an additional reference image to recover more faithful textures and structures. Existing RefSR methods mainly differ in how they establish LR--Ref correspondence and transfer reference information. Early methods perform dense patch matching in the feature domain~\cite{srntt2019,TTSR2020,zheng2017learning,yue2013landmark}, while other works introduce optical flow~\cite{crossnet2018} or deformable convolution~\cite{SSEN2020} for more flexible alignment. Later approaches further combine patch correspondences with DCN-based refinement~\cite{c2-matching2021,huang2022task,datsr2022}, or adopt coarse-to-fine matching to reduce computational cost~\cite{masa2021,AMSA2022}.

However, most RefSR methods assume that the LR and Ref images are relatively consistent in appearance and geometry. This assumption becomes unreliable in dual-lens scenarios, where cross-camera discrepancies and large parallax jointly make correspondence estimation much more challenging.

\subsection{Dual-Lens Super-Resolution}
DuSR aims to super-resolve a wide-angle LR image with the assistance of a telephoto reference image. DCSR~\cite{DCSR2021} first extends RefSR to this setting by training on synthetic data and adapting to real images through self-supervision, but it does not explicitly address large-parallax alignment. Subsequent methods improve DuSR from different perspectives. SelfDZSR~\cite{SelfDZSR2022} uses telephoto images as supervision for the center region,  while SelfDZSR++~\cite{selfdzsr++} further extends self-supervised learning to dual- and multiple-zoom observations with optical-flow-based pre-alignment and progressive multi-view fusion for misaligned supervision and additional zoom inputs. ZeDuSR~\cite{ZeDuSR2023} performs online optimization to adapt LR features to the reference, and geometric correspondence has also been explored to improve matching quality~\cite{geometry2023}. More recently, KeDuSR~\cite{2024KeDuSR} introduces kernel-free matching to alleviate camera-kernel discrepancies.

Despite these advances, existing DuSR methods still rely on implicit alignment or suboptimal geometric priors. As a result, they struggle to establish reliable correspondence under the joint challenges of large parallax, cross-camera discrepancies, and spatially varying reference similarity.

\subsection{Image Alignment}
Image alignment is essential for low-level vision tasks involving multi-frame or cross-view information fusion, such as video restoration~\cite{chan2022basicvsr++,chan2021basicvsr,liang2024vrt,liang2022rvrt,vsrt}, DuSR~\cite{2024KeDuSR,selfdzsr++,ZeDuSR2023}, and dual-camera image demoiréing~\cite{mei2025image}. Optical flow-based methods~\cite{weinzaepfel2013deepflow,Flownet,Flownet2.0,spynet,pwcnet,raft,wang2024sea,wang2025waft} and deformable convolution-based methods~\cite{dcn,dcnv2} are two representative alignment paradigms. Modern flow networks such as SPyNet~\cite{spynet}, PWC-Net~\cite{pwcnet}, and RAFT~\cite{raft} have been widely used as generic alignment priors in video restoration and related tasks~\cite{chan2022basicvsr++,chan2021basicvsr,liang2024vrt,liang2022rvrt,vsrt}.

Nevertheless, these pretrained flow models are mainly developed for monocular image pairs under assumptions such as photometric consistency and relatively smooth motion. In dual-lens SR, such assumptions are violated by resolution, color, brightness, blur, and viewpoint discrepancies between different lenses. More broadly, recent studies on few-shot learning and vision--language model adaptation have shown that pretrained representations may exhibit limited reliability, task-dependent bias, or feature redundancy when directly transferred to downstream settings with distribution shifts, motivating task-aware adaptation strategies~\cite{zhang2025reliable,zhang2026channel,zhang2026closer}. Similarly, directly applying off-the-shelf monocular flow models to dual-lens images may lead to unreliable cross-view alignment, motivating the proposed task-specific dual-lens flow learning strategy.

%% file: 3Methods.tex
\section{DuSR-RealV2 Dataset}

To support comprehensive real-world DuSR evaluation, a desirable dataset should satisfy two complementary requirements: reliable paired alignment and sufficient parallax diversity.
Well-aligned paired samples enable supervised optimization and full-reference quantitative evaluation, while large-parallax samples are necessary for assessing robustness under occlusion, depth discontinuities, and spatially inconsistent references.
However, existing real-world DuSR datasets mainly focus on constructing accurately aligned pairs, which naturally favors scenes with small parallax and simple depth layouts.
As a result, challenging cases with large viewpoint displacement and heavy occlusion are often underrepresented, since they are difficult to align at the pixel level.

To address this limitation, we construct a new 4K-resolution dataset, termed DuSR-RealV2, using the wide-angle and telephoto cameras of an iPhone 15 Pro. The two views are synchronously captured using the software Camera M\footnote{\url{https://www.camera-m.com/}}. During data acquisition, we intentionally include both distant scenes with relatively simple depth structures and challenging scenes containing nearby foreground objects, large parallax, and occlusions. Following KeDuSR~\cite{2024KeDuSR}, the captured images are processed by SIFT-based coarse alignment~\cite{lowe2004distinctive}, DeepFlow-based refinement~\cite{weinzaepfel2013deepflow}, intensity correction, and spatial cropping, resulting in aligned HR images and LR images. Since challenging scenes are not always reliably alignable at the pixel level, we further introduce a quantitative misalignment measurement to classify the dataset into paired and unpaired subsets. Note that the residual between aligned HR and LR images comes from three aspects, the brightness differences, the high-frequency differences, and the misalignment errors. Therefore, we first perform Gaussian smoothing and intensity correction
to remove the effects of the first two aspects. Then, we compute the residual error map between aligned HR and LR images, and use the percentage of pixels with large residual errors as the misalignment measure. A predefined percentage ($>0.1\%$) is used for the initial separation, followed by manual selection for borderline cases. Note that, for the Unpaired subset, we utilize the pair processed by SIFT-based coarse alignment and spatial cropping, which is beneficial for determining the overlapping region between the LR and Ref images. 
In this way, we construct the DuSR-RealV2 dataset, which contains 725 image triplets, including 493 training and 129 testing Paired samples and 103 Unpaired samples. Compared with DuSR-Real~\cite{2024KeDuSR}, DuSR-RealV2 broadens real-world DuSR evaluation beyond reliably alignable scenes by additionally retaining samples with closer foreground objects, stronger parallax, and more complex occlusion patterns.

\begin{figure}[t]
  \centering
  \includegraphics[width=1\linewidth]{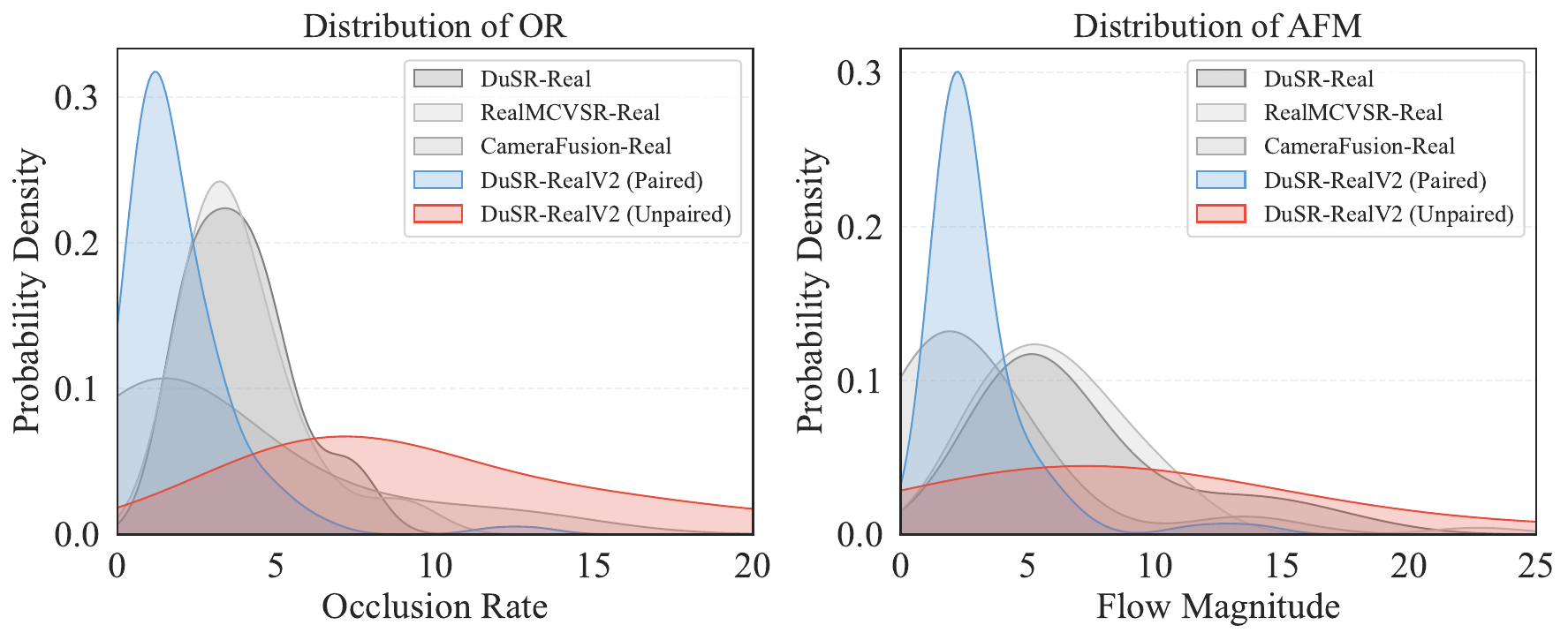}
   \caption{Distributions of Occlusion Rate (OR) and Average Flow Magnitude (AFM) across different DuSR datasets. 
   DuSR-RealV2-Paired exhibits the lowest and most compact OR distribution, indicating highly reliable paired alignment. 
   In contrast, DuSR-RealV2-Unpaired shows a broader distribution shifted toward larger AFM values, demonstrating stronger LR--Ref parallax and higher evaluation difficulty.}
   \label{fig:data_processing}
\end{figure}

We further compare different DuSR datasets using Occlusion Rate (OR) and Average Flow Magnitude (AFM), which reflect alignment reliability and parallax magnitude, respectively.
OR measures the proportion of occluded or unmatched regions between the LR and Ref views and is computed by accumulating the hole regions induced by forward and backward optical-flow warping.
A lower OR indicates fewer unreliable correspondences and thus better pixel-level alignment.
As shown in Fig.~\ref{fig:data_processing}, existing paired datasets generally exhibit relatively low OR because they mainly retain samples that can be aligned reliably.
Notably, DuSR-RealV2-Paired shows the most compact low-OR distribution, demonstrating that our paired subset provides highly reliable alignment for supervised training and full-reference evaluation.
In contrast, DuSR-RealV2-Unpaired has a broader and higher-OR distribution, indicating more severe occlusion and spatial inconsistency in challenging real-world cases.
AFM is computed as the mean magnitude of the optical flow field estimated between the LR and Ref views for cross-view alignment, and is used to characterize the degree of parallax.
Existing paired datasets, including DuSR-Real and RealMCVSR-Real, are mostly concentrated in relatively small-AFM ranges, suggesting limited viewpoint displacement due to the requirement of accurate alignment.
In comparison, DuSR-RealV2-Unpaired exhibits a much broader distribution and clearly shifts toward larger AFM values, showing that it contains substantially larger parallax than existing datasets.
Overall, the low OR of DuSR-RealV2-Paired verifies the reliability of its aligned pairs, while the large AFM of DuSR-RealV2-Unpaired demonstrates its value for evaluating robustness under severe parallax, occlusion, and complex depth variations.

\section{Method}

\begin{figure*}[t]
  \centering
  \includegraphics[width=0.95\linewidth]{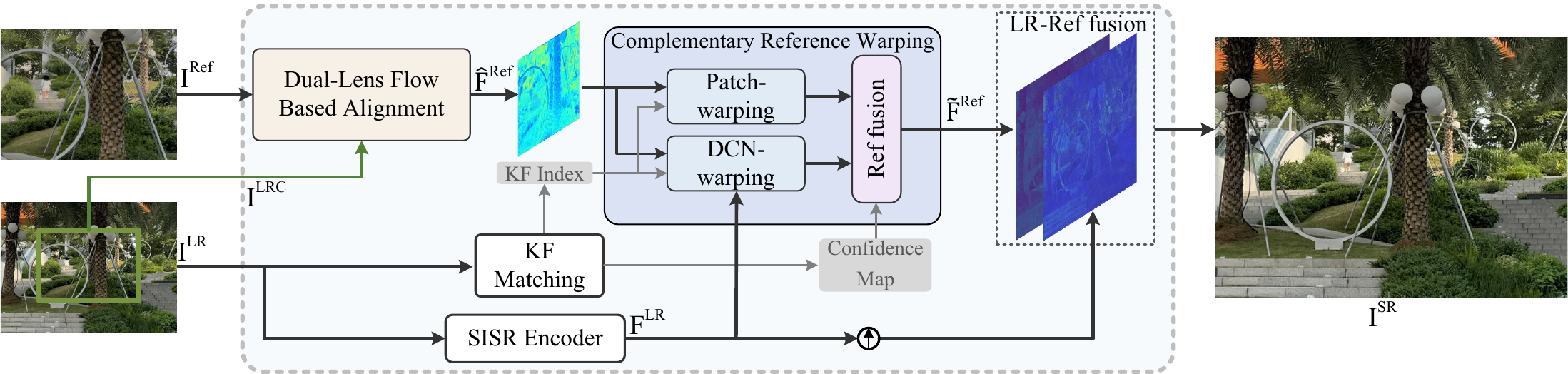}
    \caption{The framework of our DuSRFlow. The proposed dual-lens flow based alignment module first aligns the reference image $\mathbf{I}^{\mathrm{Ref}}$ to the LR center region, yielding the aligned reference feature $\hat{\mathbf{F}}^{\mathrm{Ref}}$. In parallel, kernel-free (KF) matching between $\mathbf{I}^{\mathrm{LR}}$ and $\mathbf{I}^{\mathrm{LRC}}$ produces the KF index and confidence map, while the SISR encoder extracts the LR feature $\mathbf{F}^{\mathrm{LR}}$. These features are then processed by the proposed complementary reference warping to generate the aggregated reference feature $\tilde{\mathbf{F}}^{\mathrm{Ref}}$, which is fused with the LR feature through LR--Ref fusion to produce the final SR result $\mathbf{I}^{\mathrm{SR}}$.}
   \label{fig:DuSRFlow}
\end{figure*}

\subsection{Framework Overview}

As illustrated in Fig.~\ref{fig:DuSRFlow}, the proposed DuSRFlow framework consists of three key stages: dual-lens flow based alignment, complementary reference warping, and LR--Ref fusion. First, the reference image is aligned to the center region of the LR image by the proposed dual-lens flow based alignment module, yielding the aligned reference feature $\hat{\mathbf{F}}^{\mathrm{Ref}}$.
In parallel, kernel-free (KF) matching is performed between the LR image $\mathbf{I}^{\mathrm{LR}}$ and its center region $\mathbf{I}^{\mathrm{LRC}}$ to estimate the KF index and confidence map. Since $\hat{\mathbf{F}}^{\mathrm{Ref}}$ has been aligned to the $\mathbf{I}^{\mathrm{LRC}}$ domain, these matching cues, together with the LR feature extracted by the SISR encoder, are fed into the proposed complementary reference warping module to generate the aggregated reference feature $\tilde{\mathbf{F}}^{\mathrm{Ref}}$.
Finally, the aggregated reference feature is fused with the LR feature through the LR--Ref fusion module, and the resulting representation is used to generate the final super-resolved image. The following subsections describe each component in detail.

\subsection{Dual-Lens Flow Based Alignment}

As an important component of the proposed DuSRFlow framework, we design a Dual-Lens Flow Based Alignment module to provide reliable geometric priors for subsequent reference transfer. At the core of this module is DuFlowNet, a dedicated dual-lens flow estimator trained with the proposed real--synthetic data fusion strategy.

\begin{figure}[t]
  \centering
  \includegraphics[width=0.8\linewidth]{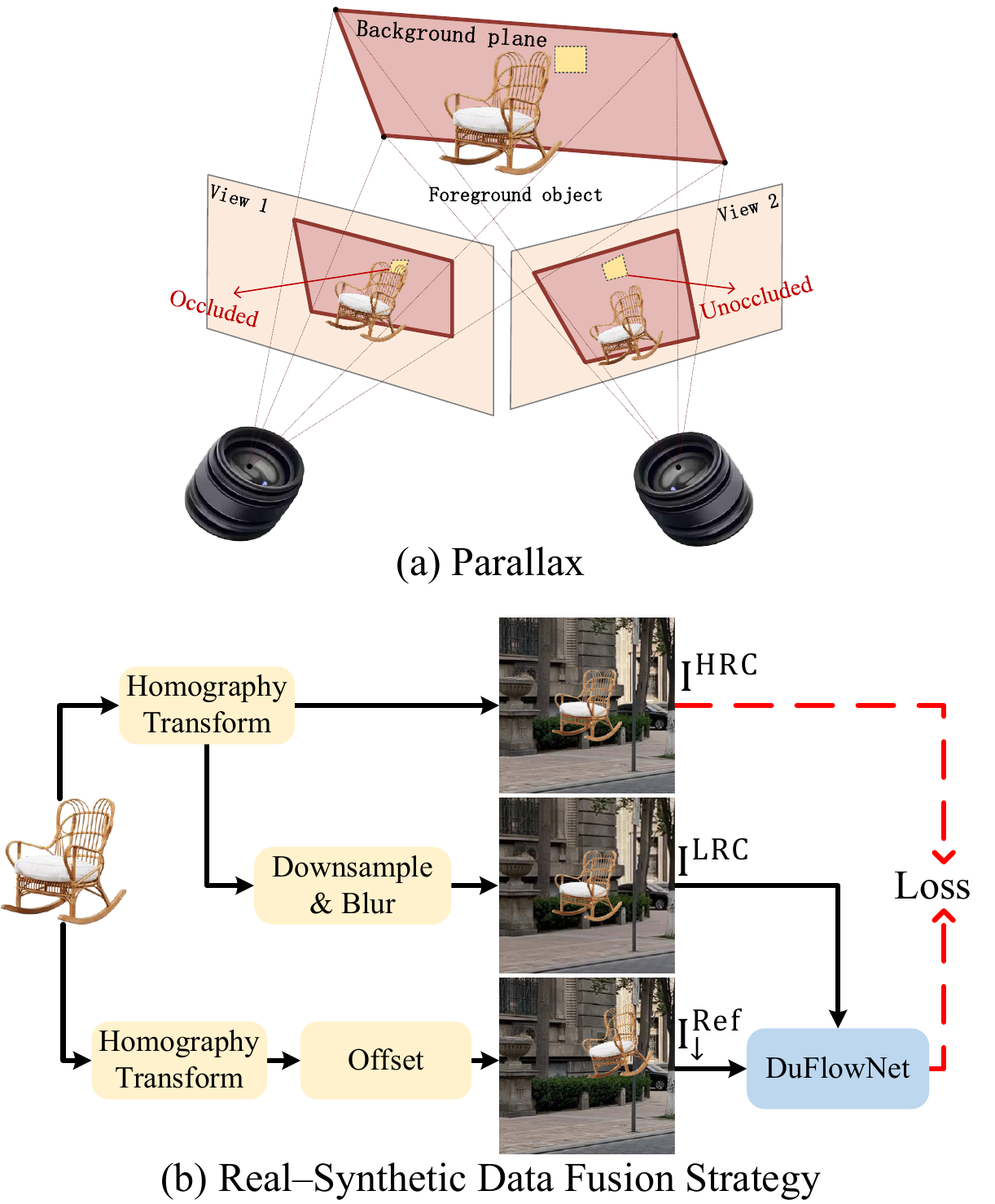}
   \caption{(a) Illustration of dual-lens parallax and view-dependent occlusion. A foreground object can exhibit different projections and occlude different background regions across views. (b) The proposed real--synthetic data fusion strategy. A synthetic foreground is composited onto real dual-lens image pairs, where different homography transformations are applied to the reference branch and the LR/HR-center branches to generate controllable parallax while preserving realistic dual-lens imaging characteristics.}
   \label{fig:flownet}
\end{figure}

\subsubsection{Real–Synthetic Data Fusion Strategy}
Existing optical flow networks, such as SPyNet~\cite{spynet}, RAFT~\cite{raft}, and PWC-Net~\cite{pwcnet}, are typically pretrained on monocular motion datasets such as FlyingChairs~\cite{Flownet} and later adapted to video restoration tasks. However, these models perform poorly in dual-lens scenarios, where cross-camera discrepancies in resolution, color, and brightness introduce substantial domain gaps. Under large parallax, they often fail to provide reliable alignment priors for dual-lens correspondence.

Nevertheless, existing dual-lens data alone are still insufficient for training a dual-lens flow model that is robust to large parallax. In DuSR datasets, dual-lens pairs can be roughly divided into two types: unpaired samples with large parallax but no reliable pixel-level supervision, and paired samples with reliable LR--HR alignment but limited parallax. Thus, neither type alone is sufficient to train a robust dual-lens flow model.

To address this issue, we propose a real--synthetic data fusion strategy for training a dedicated dual-lens flow network. Our goal is to construct training samples with controllable parallax while preserving realistic cross-camera discrepancies. As illustrated in Fig.~5(a), depth differences between the foreground and background can lead to cross-view displacement and view-dependent occlusion. Based on this observation, we retain the paired real data as the background to preserve authentic dual-lens imaging characteristics, while introducing composited foreground objects with different transformations across views to simulate controllable parallax and view-dependent occlusion.

Specifically, as shown in Fig.~\ref{fig:flownet}(b), a virtual foreground undergoes two random homography transformations: one is applied to the reference branch, and the other is applied to both the LR-center image $\mathbf{I}^{\mathrm{LRC}}$ and the corresponding HR-center image $\mathbf{I}^{\mathrm{HRC}}$. In addition, a relative offset is introduced between the composited foreground regions across branches to simulate view-dependent occlusion inconsistency caused by depth variation. The composited foreground in the LR branch is further blurred to mimic the resolution gap between LR and HR. In this way, we obtain realistic training triplets with controllable parallax while preserving real background structure and cross-camera discrepancies.

\subsubsection{DuFlowNet for Dual-Lens Alignment}
As the core of the proposed Dual-Lens Flow Based Alignment module, \textbf{DuFlowNet} estimates cross-view correspondence from the reference branch to the LR-center domain. DuFlowNet adopts an efficient U-Net-style encoder--decoder architecture following~\cite{HDRFlow}. Specifically, the downsampled reference image $\mathbf{I}^{\mathrm{Ref}}_{\downarrow}$ and the LR-center image $\mathbf{I}^{\mathrm{LRC}}$ are concatenated and fed into DuFlowNet to predict the optical flow field $\mathbf{f}$.

The estimated flow is first used to warp the downsampled reference image:

\begin{equation}
\mathbf{I}^{\text{Ref}}_{\text{W}}=
\omega\!\left(\mathbf{I}^{\text{Ref}}_{\downarrow},\mathbf{f}\right),
\end{equation}
where $\omega(\cdot)$ denotes backward warping parameterized by $\mathbf{f}$.
Instead of using ground-truth flow annotations, we supervise the warped result with the downsampled HR-center image $\mathbf{I}^{\mathrm{HRC}}_{\downarrow}$, which shares the target view with $\mathbf{I}^{\mathrm{LRC}}$ and the same telephoto-camera response as $\mathbf{I}^{\mathrm{Ref}}$, providing a photometrically consistent target for learning cross-view correspondence. Note that DuFlowNet predicts only a two-dimensional flow field for spatial warping and does not perform appearance transformations; therefore, it learns geometric alignment rather than a mixture of alignment and appearance correction.

The estimated flow is further used for feature-level alignment in the SR task. Since direct flow-based warping is sensitive to local errors, we employ a flow-guided DCN to refine the warped reference features. Guided by the upsampled flow, the module aligns the reference features to the LR-center domain and produces the aligned reference feature $\hat{\mathbf{F}}^{\mathrm{Ref}}$.

\begin{figure}[t]
  \centering
  \includegraphics[width=1\linewidth]{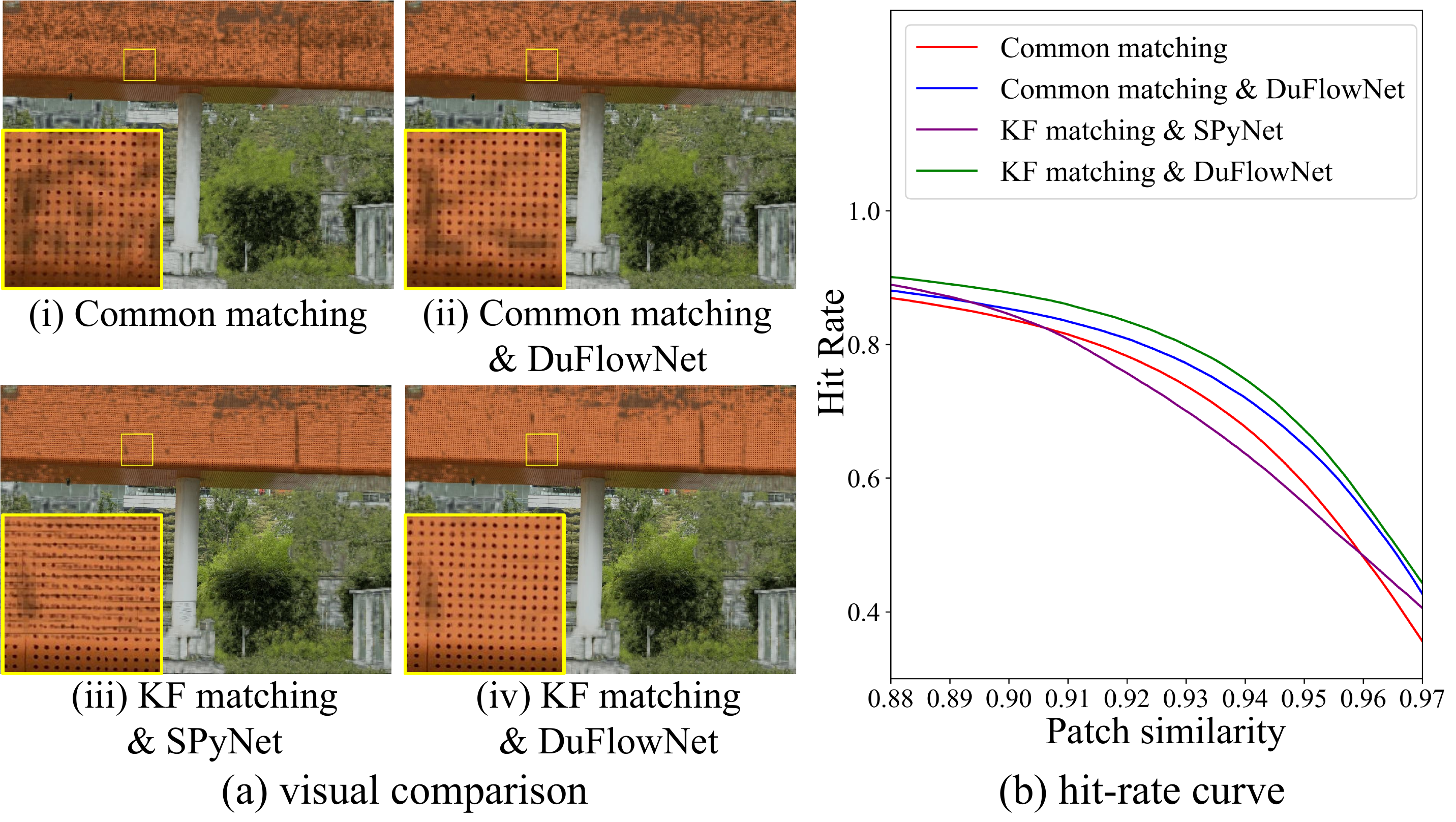}
   \caption{A toy experiment to evaluate DuFlowNet. (a) HR image reconstruction with pixel-domain patch matching using different matching methods. 
(b) Patch similarity and hit rate curves with different matching methods.}
   \label{fig:alignment_validation}
\end{figure}

\subsubsection{Alignment Effectiveness Validation}

To verify that DuFlowNet provides effective geometric priors for subsequent reference matching, we conduct a toy experiment in which an HR proxy is reconstructed by directly assembling matched reference patches in the pixel domain. As shown in Fig.~\ref{fig:alignment_validation}(a), four settings are compared: (i) Common matching, which directly performs patch matching between LR and Ref; (ii) Common matching \& DuFlowNet, which first aligns the reference using DuFlowNet before common matching; (iii) KF matching \& SPyNet, which combines KF matching with SPyNet-based alignment; and (iv) KF matching \& DuFlowNet, which combines KF matching with the proposed alignment.

Fig.~\ref{fig:alignment_validation}(a) shows that, compared with Common matching, Common matching \& DuFlowNet exhibits fewer blocky artifacts, indicating that DuFlowNet provides better alignment priors for subsequent patch matching. Similarly, compared with KF matching \& SPyNet, KF matching \& DuFlowNet yields a more consistent reconstruction with fewer line-like and structural artifacts.

We further analyze the statistical behavior of patch correspondence in Fig.~\ref{fig:alignment_validation}(b). Patch similarity measures the similarity between each matched reference patch and its corresponding HR patch, while hit rate denotes the proportion of matched patches whose similarity exceeds a given threshold. Under the same patch-similarity threshold, DuFlowNet consistently achieves a higher hit rate than the compared alternatives, confirming that it provides more reliable geometric priors for subsequent matching.

\subsection{Complementary Reference Warping}

Existing RefSR methods usually focus on local reference transfer, but overlook that unreliable local reference features may suppress the utilization of reliable reference cues even in well-matched regions. To address this issue, we introduce a Complementary Reference Warping module that coordinates patch-based warping and KF-guided DCN-based warping (KF-DCN warping) to construct a more reliable reference representation before LR--Ref fusion.

\subsubsection{Kernel-Free Matching}
Patch matching is beneficial for regions with high-similarity \cite{2015Image}. 
Following~\cite{srntt2019,TTSR2020}, we perform kernel-free patch matching in the VGG feature space~\cite{vgg} between the LR image $\mathbf{I}^{\mathrm{LR}}$ and its center region $\mathbf{I}^{\mathrm{LRC}}$. Specifically, the corresponding feature maps $\mathbf{F}^{\mathrm{LR}}_{\mathrm{vgg}}$ and $\mathbf{F}^{\mathrm{LRC}}_{\mathrm{vgg}}$ are divided into $3\times3$ patches. For each LR patch $\mathbf{P}^{\mathrm{LR}}_i$, we identify the most similar patch $\mathbf{P}^{\mathrm{LRC}}_j$ by cosine similarity $S_{i,j}$, and obtain the matching index $M_i$ and confidence score $C_i$ as
\begin{equation}\label{index}
M_i=\arg\max_j S_{i,j}, \qquad C_i=\max_j S_{i,j}.
\end{equation}
The resulting KF index $M_i$ and confidence score $C_i$ are used for subsequent reference warping and reference fusion.

\subsubsection{Patch-Based Warping}

Since the aligned reference feature $\hat{\mathbf{F}}^{\mathrm{Ref}}$ has already been mapped to the $\mathbf{I}^{\mathrm{LRC}}$ domain by DuFlowNet, the KF index $M_i$ can be directly reused for reference transfer. The patch-warped reference feature is thus constructed as
\begin{equation}
\tilde{\mathbf{F}}^{\mathrm{Ref}}_{\mathrm{KP},P_i}
=
\hat{\mathbf{F}}^{\mathrm{Ref}}_{M_i},
\end{equation}
where $\tilde{\mathbf{F}}^{\mathrm{Ref}}_{\mathrm{KP},P_i}$ denotes the warped reference feature associated with patch $P_i$.

\subsubsection{KF-Guided DCN-Based Warping}

Although patch warping preserves structurally faithful transfer in highly similar regions, it becomes less reliable in weakly correlated areas such as corners and occlusion boundaries. To improve reference transfer in these regions, we introduce a KF-DCN warping branch, denoted as \textit{KF-DCN}. Instead of relying on rigid patch retrieval alone, this branch uses the KF index as a coarse correspondence anchor and performs deformable aggregation around the matched location.

For each position $P_i$, the warped reference feature $\tilde{\mathbf{F}}^{\mathrm{Ref}}_{\mathrm{KD},P_i}$ is computed as
\begin{equation}
\tilde{\mathbf{F}}^{\mathrm{Ref}}_{\mathrm{KD},P_i}
=
\sum_{k=1}^{K}
w_k \cdot
\hat{\mathbf{F}}^{\mathrm{Ref}}\!\left(M_i+p_k+\Delta p_k\right)
\cdot \Delta m_k,
\end{equation}
where $K$ is the number of sampling locations, $w_k$ denotes the shared kernel weight, $p_k$ is the predefined sampling offset, and $\Delta p_k$ and $\Delta m_k$ are the learnable deformable offsets and modulation masks, respectively. These offsets and masks are predicted by a lightweight convolutional block conditioned on the LR feature, the patch-warped reference feature, and the aligned reference feature.

\subsubsection{Complementary Reference Fusion}

The above two branches produce two warped reference features for the same LR input: the patch-warped feature $\tilde{\mathbf{F}}^{\mathrm{Ref}}_{\mathrm{KP}}$ and the KF-DCN-warped feature $\tilde{\mathbf{F}}^{\mathrm{Ref}}_{\mathrm{KD}}$. A straightforward solution is to directly concatenate them or pass both branches to the downstream LR--Ref fusion module. However, such a design cannot adequately handle their spatially varying reliability. More importantly, once unreliable local reference features are mixed into the reference branch, they may reduce its overall reliability and consequently suppress the utilization of reliable reference cues even in well-matched regions.

To alleviate this effect, we first perform complementary fusion within the reference branch before LR--Ref fusion. Specifically, we transform the confidence map $\mathbf{C}$ into a spatially adaptive modulation map $\hat{\alpha}$ by a lightweight mapping function $g(\cdot)$, which consists of two convolutional layers, each followed by a LeakyReLU activation:
\begin{equation}
\hat{\alpha}=g(\mathbf{C}).
\end{equation}
The final aggregated reference feature is then computed as
\begin{equation}
\tilde{\mathbf{F}}^{\mathrm{Ref}}
=
\hat{\alpha}\odot\tilde{\mathbf{F}}^{\mathrm{Ref}}_{\mathrm{KP}}
+
(1-\hat{\alpha})\odot\tilde{\mathbf{F}}^{\mathrm{Ref}}_{\mathrm{KD}},
\end{equation}
where $\odot$ denotes element-wise multiplication.

This design emphasizes patch warping in highly similar regions while relying more on KF-DCN warping in weakly correlated or occluded areas. Therefore, the proposed fusion is not a simple combination of two branches, but a confidence-aware coordination mechanism that constructs a more reliable reference representation before downstream LR--Ref fusion.

\begin{figure}[t]
    \centering
    \includegraphics[width=0.8\linewidth]
    {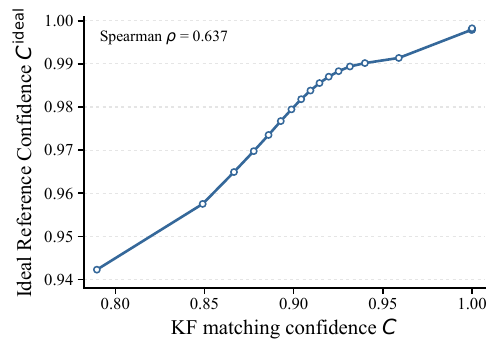}
    \caption{Validation of KF confidence as a proxy for cross-camera reference-transfer reliability. All valid patches are divided into 20 equal-count bins according to KF confidence $C$, and each point reports the mean $C$ and mean ideal reference confidence $C^{\mathrm{ideal}}$ within each bin. The Spearman correlation over all patches is $0.637$.
}
    \label{fig:confidence_reliability}
\end{figure}

To verify whether KF confidence reflects cross-camera transfer reliability, we conduct a patch-level analysis on the DuSR-RealV2-Paired test set. For each LR patch $P^{LR}_i$, KF matching provides confidence $C_i$ and index $M_i$. We retrieve $P^{\mathrm{AlignedRef}}_{M_i}$ from the DuFlowNet-aligned reference and define the ideal reference confidence using the corresponding HR patch as { \begin{equation} C_i^{\mathrm{ideal}} = \cos\left( P^{\mathrm{AlignedRef}}_{M_i}, P^{HR}_i \right). \end{equation} } While $C_i^{\mathrm{ideal}}$ directly measures the reliability of the transferred reference content, it requires the HR image and is unavailable during inference. We therefore examine whether the inference-available KF confidence $C_i$ can serve as its proxy. As shown in Fig.~\ref{fig:confidence_reliability}, after dividing all valid patches into 20 equal-count bins according to $C_i$, the mean $C_i^{\mathrm{ideal}}$ exhibits a clear increasing trend with $C_i$. The Spearman correlation over all valid patches reaches $0.637$, demonstrating a clear positive relationship between KF confidence and ideal reference confidence. These results support using KF confidence to indicate cross-camera reference-transfer reliability.

\subsection{LR--Ref Fusion}

Following existing RefSR practice~\cite{huang2022task}, we employ a standard SISR encoder to extract the LR feature $\mathbf{F}^{\text{LR}}$ and upsample it to the same resolution as the aggregated reference feature $\tilde{\mathbf{F}}^{\text{Ref}}$. The two features are then fused by an adaptive fusion module with spatial and channel attention, and the fused representation is passed to a convolutional reconstruction head to generate the final SR result $\mathbf{I}^{\text{SR}}$.

%% file: 4Experiment.tex
\section{Experiments}
Detailed training settings and loss functions are provided in the supplementary material.
\begin{table}[t]
  \centering
  \caption{
  Quantitative comparisons on DuSR-RealV2-Paired. Bold and underlined indicate best and second-best, respectively. -\(\ell\) denotes training with only reconstruction loss.
  }
    \input{tables/1}

  \label{tab:DuSR-Real-V2-paired-SR}
\end{table}

\begin{table}[t]
  \centering
  \caption{ Quantitative comparisons on DuSR-Real. }
    \input{tables/2}

  \label{tab:DuSR-Real}
\end{table}

\begin{table}[t]
  \centering
  \caption{Quantitative comparisons on DuSR-RealV2-Unpaired. 
Since ground-truth HR images are unavailable, we employ no-reference metrics for objective evaluation. }
    \input{tables/5}
  \label{tab:DuSR-Real-V2-unpaired-SR}
\end{table}

\begin{table}[t]
\centering
{
\caption{Comparison of optical flow architectures on DuSR-RealV2 using official pretrained weights and the proposed real--synthetic data fusion strategy. Values in parentheses denote changes relative to the corresponding pretrained models. Latency is measured for estimating one $512\times960$ flow field on an NVIDIA RTX 3090 GPU.}
\label{tab:DuSR-RealV2-paired-flow-main}
\input{tables/12}
}
  
\end{table}

\subsection{Quantitative Comparison of SR Methods}

To evaluate the effectiveness of the proposed DuSRFlow, we compare it with three categories of SR methods: SISR methods, including RCAN~\cite{rcan2018}, SwinIR~\cite{swinir2021}, ESRGAN~\cite{esrgan2018}, and BSRGAN~\cite{bsrgan2021}; RefSR methods, including TTSR~\cite{TTSR2020}, MASA~\cite{masa2021}, and DATSR~\cite{datsr2022}; and DuSR methods, including DCSR~\cite{DCSR2021}, SelfDZSR~\cite{SelfDZSR2022}, ZeDuSR~\cite{ZeDuSR2023}, and KeDuSR~\cite{2024KeDuSR}. The SISR baselines are trained on LR--HR pairs, whereas the RefSR baselines are trained on LR--Ref--HR triplets.

We evaluate all methods on four real-world DuSR datasets, including DuSR-Real~\cite{2024KeDuSR}, our newly constructed DuSR-RealV2, RealMCVSR-Real, and CameraFusion-Real. In the main paper, we report the results on DuSR-Real and DuSR-RealV2, while the results on RealMCVSR-Real and CameraFusion-Real are provided in the supplementary material.
Results on DuSR-Real are reported in Table~\ref{tab:DuSR-Real}.
For DuSR-RealV2, we separately evaluate the paired and unpaired subsets, with results shown in Tables~\ref{tab:DuSR-Real-V2-paired-SR} and~\ref{tab:DuSR-Real-V2-unpaired-SR}, respectively.
Since the DuSR-RealV2-Unpaired subset does not provide well-aligned HR images, we evaluate it using no-reference metrics, including NRQM~\cite{NRQM}, NIQE~\cite{niqe}, PI~\cite{PI}, BRISQUE~\cite{BRISQUE}, MUSIQ~\cite{ke2021musiq}, and MANIQA~\cite{yang2022maniqa}.
For the paired benchmarks, i.e., DuSR-Real and DuSR-RealV2-Paired, we report full-reference metrics over three spatial scopes: \textit{full-image}, computed over the entire image; \textit{center-image}, computed over the overlapping field of view; and \textit{corner-image}, computed over the remaining non-overlapping regions.
Methods trained using reconstruction loss only are marked with the suffix ``-$\ell$'', whereas the others use all available losses.

As shown in Tables~\ref{tab:DuSR-Real-V2-paired-SR} and~\ref{tab:DuSR-Real}, DuSRFlow achieves the best overall performance on the paired benchmarks in terms of PSNR, SSIM~\cite{SSIM}, and LPIPS~\cite{Lpips}. On the more challenging DuSR-RealV2-Unpaired, it also obtains the best results across all no-reference metrics, showing stronger robustness to severe parallax, occlusion, and spatially inconsistent references.

\begin{figure*}
  \centering

  \includegraphics[width=0.9\linewidth]{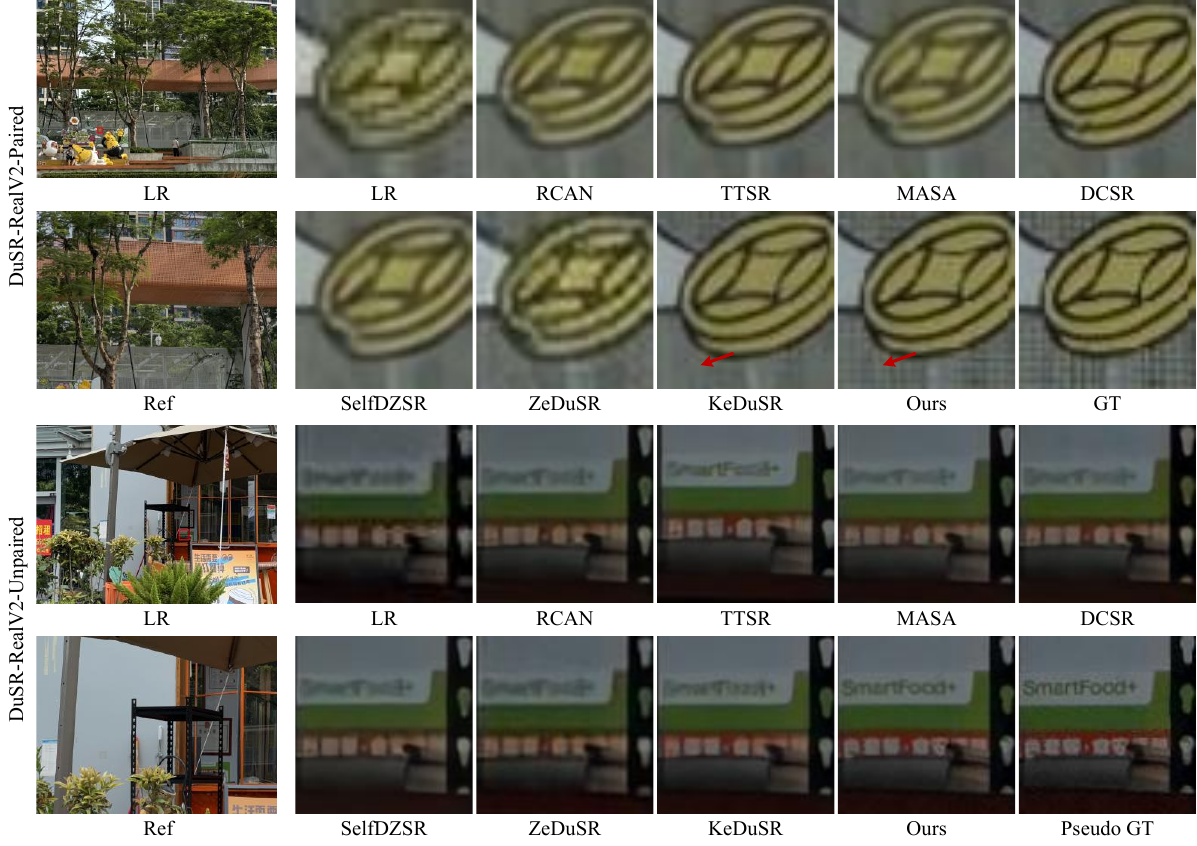}
   \caption{Visual comparisons on the DuSR-RealV2 dataset. The presented results are obtained with only the reconstruction loss. Zoom in for better observation. The Pseudo ground truth of the unpaired subset is provided for reference only.}
   \label{sec:visual}
\end{figure*}

Fig.~\ref{sec:visual} shows the results on the DuSR-RealV2 dataset. The reconstructed results demonstrate that our method achieves the best visual quality, including more faithful texture details and closer resemblance to the HR images. Our approach effectively leverages the texture information from the Ref images to recover details that are not present in the LR images. More visual comparisons on DuSR-RealV2-Paired, DuSR-RealV2-Unpaired, RealMCVSR-Real, DuSR-Real, and CameraFusion-Real are provided in the supplementary material.

DuSRFlow also shows strong generalization and competitive efficiency. It achieves the best cross-dataset generalization performance and the second-fastest inference speed among the compared RefSR methods. Detailed results are provided in the supplementary material.

\begin{table}
  \centering
\caption{Ablation study on DuFlowNet Prior and KF-DCN}
\input{tables/8}
  \label{tab:Ablation SR}
\end{table}

\begin{table}
  \centering
\caption{Ablation study on the Complementary Reference Warping}
\input{tables/9}

  \label{tab:Ablation on Complementary Reference Warping}
\end{table}

\subsection{Quantitative Comparison of Flow Methods}
To evaluate the alignment quality of the proposed DuFlowNet, we compare it with six representative optical flow models, namely SPyNet~\cite{spynet}, PWC-Net~\cite{pwcnet}, RAFT~\cite{raft}, GMFlow~\cite{xu2022gmflow}, SEA-RAFT~\cite{wang2024sea}, and WAFT~\cite{wang2025waft}. For all compared methods, we use their officially released pretrained weights to align the reference image toward the LR-center view. In contrast, DuFlowNet is trained with the proposed real--synthetic data fusion strategy, which is specifically designed to mimic dual-lens parallax and improve robustness under large cross-camera discrepancies.

Since there is no real dual-lens dataset with ground-truth flow, endpoint error cannot be used for evaluation. Instead, we assess how well the aligned reference matches the target center view, which directly reflects its usefulness for subsequent reference-based reconstruction. Specifically, on the DuSR-RealV2-Paired subset, we compute PSNR and SSIM between the aligned reference image and the HR-center image. On the DuSR-RealV2-Unpaired subset, where well-aligned HR ground truth is unavailable, we compute PSNR and SSIM between the aligned reference image and the LR-center image as a proxy for cross-view alignment consistency.
As shown in Table~\ref{tab:DuSR-RealV2-paired-flow-main}, directly applying officially pretrained optical flow models yields limited alignment accuracy on the dual-lens data. After adaptation with the proposed real--synthetic data fusion strategy, most evaluated architectures achieve consistent improvements on both the Paired and Unpaired subsets, demonstrating the effectiveness and generality of the proposed training strategy across different flow architectures. GMFlow shows limited adaptation, likely because its global correlation and position-weighted correspondence formulation is less compatible with the indirect pixel-level supervision used in our training strategy. Although some larger architectures achieve higher absolute alignment accuracy, DuFlowNet provides competitive performance with only 0.65 M parameters and 5 ms latency, offering a favorable accuracy--efficiency trade-off for dual-lens alignment.

\begin{figure}[t]
  \centering

  \includegraphics[width=1\linewidth]{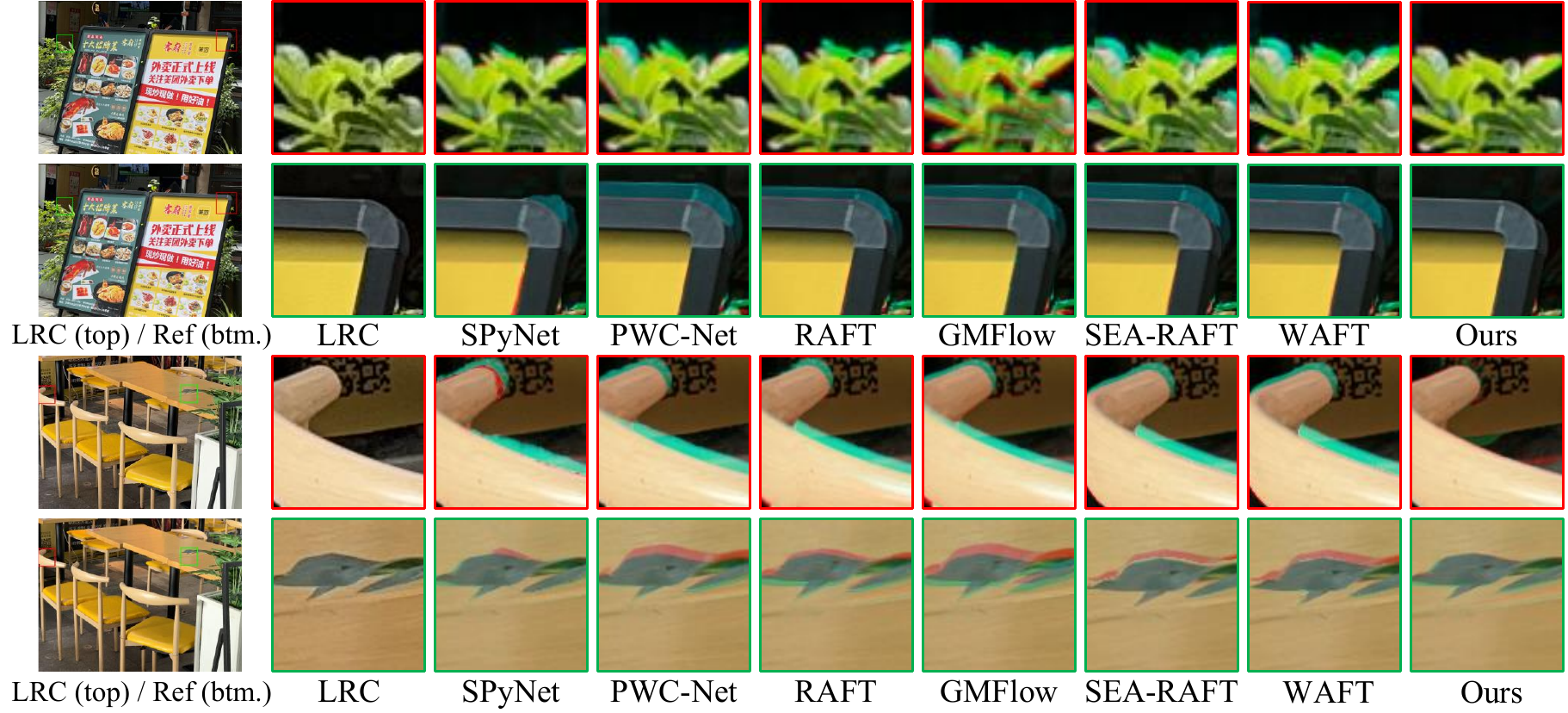}

   \caption{Visual comparison of optical flow based alignment results on the DuSR-RealV2-Unpaired subset. The artifacts in the figure indicate the displacement differences between the aligned results and the LR images.}
   \label{sec:visual4}
\end{figure}

We further provide qualitative comparisons in Fig.~\ref{sec:visual4}. To make the residual misalignment more visible, we replace the red channel of the aligned reference image with that of the LR-center image, so that structural discrepancies are highlighted as color artifacts. DuFlowNet produces the cleanest alignment with the fewest visible artifacts, especially under severe parallax and occlusion.

\subsection{Ablation Study of DuSRFlow}
To validate the effectiveness of the proposed DuSRFlow, we conduct ablation studies on the DuSR-RealV2-Paired dataset. We first analyze the contributions of the dual-lens flow prior and KF-DCN, and then study the effectiveness of the proposed Complementary Reference Warping module.

\subsubsection{Ablation on DuFlowNet Prior and KF-DCN}

Table~\ref{tab:Ablation SR} reports the ablation results of the key modules. Variant~\#1 serves as the baseline, where both the dual-lens flow prior and the KF-DCN module are removed. Variants~\#2 and \#3 introduce SPyNet and DuFlowNet as the flow prior, respectively. Compared with the flow-free baseline, explicit flow guidance improves reference alignment, while DuFlowNet brings more consistent gains, especially in the center region. This confirms that a dedicated dual-lens flow prior is more suitable than generic monocular flow for highly correlated regions.

We then study reference aggregation in low-similarity regions. Variant~\#4 adds DCN-based aggregation guided by direct LR--Ref matching indices, which improves corner-region reconstruction but remains affected by cross-camera discrepancies. Variant~\#5 further replaces the direct matching indices with the proposed KF indices to guide deformable aggregation. It achieves the best overall performance, demonstrating that DuFlowNet and KF-guided DCN are both essential for robust reference transfer in DuSR.

\subsubsection{Ablation on Complementary Reference Warping}

Table~\ref{tab:Ablation on Complementary Reference Warping} compares different ways of combining patch warping and KF-DCN warping. Variants~\#1 and \#2 keep only one branch, respectively. The results show clear complementarity: patch warping better preserves center details with high LR--Ref similarity, while KF-DCN is more effective in low-similarity corner regions.

Variants~\#3 and \#4 examine two naive fusion strategies. Region-wise stitching assigns patch warping to the center and KF-DCN to the corners, but its rigid partition cannot adapt to spatially varying reliability. Channel-concat directly passes the concatenated reference features to the subsequent LR--Ref fusion module, where channel and spatial attention implicitly learn their fusion weights. However, without explicit spatial reliability cues, such implicit fusion cannot fully preserve the regional advantages of the two branches, particularly the advantage of KF-DCN warping in low-similarity corner regions.

Variant~\#5 uniformly aggregates the patch-warped and KF-DCN-warped reference features and applies confidence only at the subsequent LR--Ref fusion stage. Its lower center-region performance indicates that, once reference responses with different reliability are uniformly mixed, later confidence modulation cannot fully prevent unreliable responses from suppressing reliable cues. Our method instead performs confidence-aware aggregation before LR--Ref fusion, thereby better preserving reliable reference cues and preventing unreliable responses from suppressing their utilization.

Variant~\#6 adopts the proposed Complementary Reference Warping and achieves the best overall performance. It improves not only the corner region over patch-only warping, but also the center region over KF-DCN-only warping, demonstrating that our design converts branch complementarity into cooperative gains rather than a simple trade-off.

\begin{figure}
  \centering
  \includegraphics[width=0.9\linewidth]{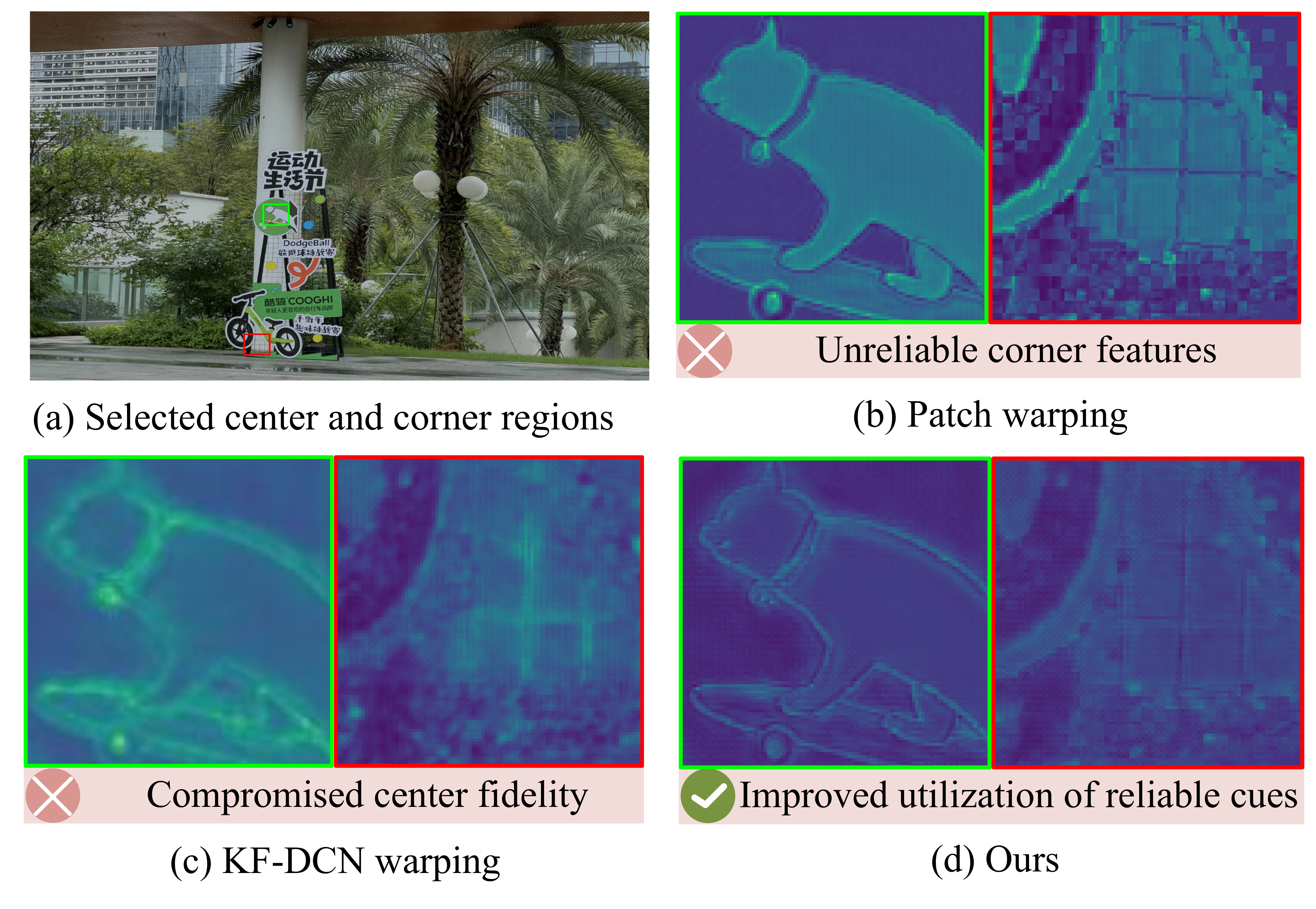}
   \caption{Visualization of reference feature aggregation results. Green and red boxes indicate center and corner regions, respectively.} 
   \label{sec:ablation ref-fusion}
\end{figure}

In Fig.~\ref{sec:ablation ref-fusion}, we visualize reference feature aggregation results
of different variants. Green and red boxes correspond to
crops from center and corner regions, respectively. Variant
(b) produces finer textures in the center but exhibits distorted,
noisy details in corners due to patch-warping limitations, while
Variant (c) shows blurrier center boundaries. Variant (d) yields spatially coherent, high-quality features via the proposed complementary reference fusion.

We verify the real--synthetic data fusion strategy in the supplementary material. Removing parallax simulation notably degrades alignment on the large-parallax unpaired subset, confirming its importance for robust dual-lens flow estimation.

\section{Conclusion and Discussion}

In this work, we presented DuSRFlow for real-world dual-lens super-resolution. We proposed DuFlowNet with a real--synthetic data fusion strategy to establish reliable cross-view correspondence between the telephoto reference and the wide-angle LR input. Building on the aligned reference features, we introduced a Complementary Reference Warping module, where patch warping preserves reliable transfer in highly similar regions, KF-guided deformable warping improves aggregation in weakly correlated regions, and complementary fusion adaptively coordinates the two branches. Extensive experiments on four benchmark datasets show that DuSRFlow consistently outperforms 11 state-of-the-art methods and generalizes well to challenging real-world scenarios.

\begin{figure}[t]
    \centering
    \includegraphics[width=0.95\linewidth]{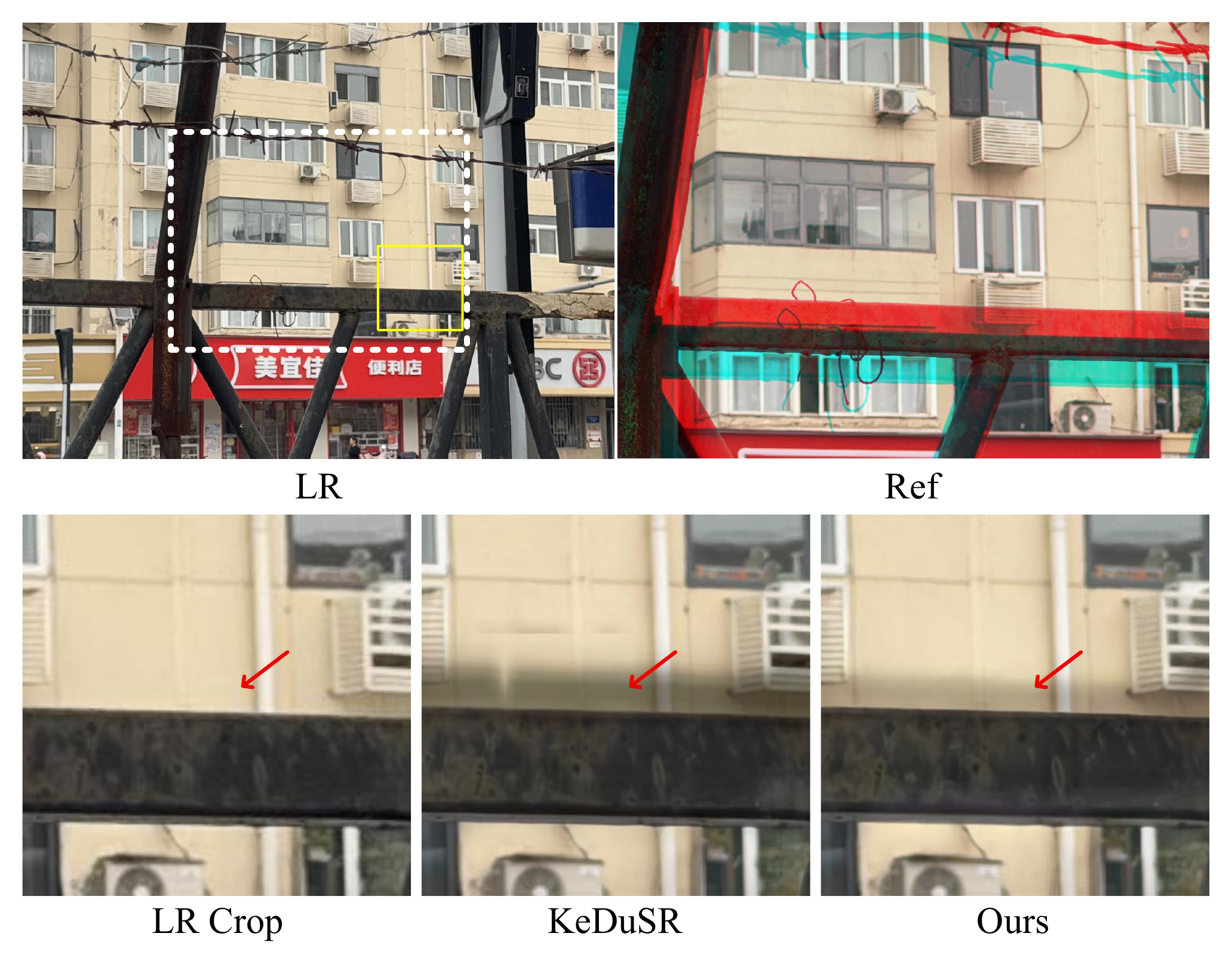}
\caption{Failure case under extremely large parallax. The white dashed box indicates the center region, and the yellow box marks the enlarged area. The color artifacts in the Ref image visualize the large LR--Ref displacement. Under such extreme parallax, DuSRFlow still exhibits residual fusion artifacts, although fewer than KeDuSR.}
    \label{fig:failure_case}
\end{figure}

\section{Limitations}

Although the proposed real--synthetic data fusion strategy effectively improves alignment under large parallax, the current foreground simulation mainly provides controllable parallax and occlusion augmentation and may not fully cover the extreme geometric variations encountered in real scenes. As shown in Fig.~\ref{fig:failure_case}, under extremely large cross-view displacement, residual alignment errors may still occur and propagate to reference transfer, introducing visible fusion artifacts in the reconstructed result. Future work will explore more flexible foreground simulation and more robust cross-view correspondence modeling for such challenging cases.

%% file: tables/1.tex
  \resizebox{\linewidth}{!}{

\begin{tabular}{lccc}
\toprule
\multirow{2}{*}{Method}  
& Full-Image        
& Center-Image 
& Corner-Image
\\
\cmidrule(lr){2-4}
& PSNR\(\uparrow\)/SSIM\(\uparrow\)/LPIPS\(\downarrow\)    
& PSNR/SSIM   
& PSNR/SSIM    
\\ \midrule

RCAN-\(\ell\)                   
& 27.16 / 0.8256 / 0.266
& 27.77 / 0.8298
& 27.02 / 0.8242 \\

RCAN                    
& 26.25 / 0.7866 / 0.145
& 26.85 / 0.7860
& 26.12 / 0.7868 \\

SwinIR-\(\ell\)                  
& 26.76 / 0.8156 / 0.277
& 27.39 / 0.8199
& 26.62 / 0.8142 \\

SwinIR                  
& 26.50 / 0.7875 / 0.151
& 27.04 / 0.7867
& 26.38 / 0.7878 \\

ESRGAN                  
& 26.39 / 0.7861 / 0.145
& 26.92 / 0.7867
& 26.27 / 0.7858 \\

BSRGAN                  
& 26.12 / 0.7844 / 0.141
& 26.56 / 0.7848 
& 26.03 / 0.7842 \\ 
\midrule

TTSR-\(\ell\)                    
& 27.93 / 0.8357 / 0.248
& 30.16 / 0.8533
& 27.43 / 0.8298 \\

TTSR                   
& 27.39 / 0.8002 / 0.124
& 29.52 / 0.8281
& 26.90 / 0.7908 \\

MASA-\(\ell\)                    
& 27.30 / 0.8218 / 0.274
& 29.52 / 0.8354
& 26.81 / 0.8172 \\

MASA                   
& 26.76 / 0.7769 / 0.151
& 28.58 / 0.7945 
& 26.34 / 0.7709 \\

DATSR-\(\ell\)                   
& 27.08 / 0.8234 /  0.273
& 27.77 / 0.8269
& 26.91 / 0.8222 \\ 

DATSR                   
& 26.67 / 0.7864 /  0.148
& 27.29 / 0.7887
& 26.52 / 0.7857 \\  
\midrule

DCSR-\(\ell\)                    
& 27.50 / 0.8322 / 0.252 
& 29.85 / 0.8536
& 26.97 / 0.8250 \\

DCSR                
& 27.05 / 0.7922 / 0.133
& 29.08 / 0.8181
& 26.58 / 0.7834 \\

SelfDZSR-\(\ell\)                
& 27.69 / 0.8310 / 0.264
& 29.07 / 0.8370 
& 27.36 / 0.8289 \\

SelfDZSR            
& 27.16 / 0.7935 / 0.137
& 28.38 / 0.7972
& 26.86 / 0.7922 \\

ZeDuSR-\(\ell\)            
& 26.66 / 0.7911 / 0.276
& 28.00 / 0.8020
& 26.34 / 0.7874 \\

ZeDuSR          
& 26.15 / 0.7753 / 0.229
& 27.46 / 0.7847
& 25.84 / 0.7722 \\

KeDuSR-\(\ell\)               
&  29.05 /  \underline{0.8609} / 0.207
& 34.79 / 0.9320
&  \underline{28.13} /  \underline{0.8368} \\

KeDuSR           
& 28.45 / 0.8269 /  \underline{0.103}
& 33.94 / 0.9175
& 27.56 / 0.7964 \\
\midrule

Ours-\(\ell\)               
& \textbf{29.76} / \textbf{0.8711} / 0.193
& \textbf{38.43} / \textbf{0.9683} 
& \textbf{28.68} / \textbf{0.8383} \\

Ours           
&  \underline{29.11} / 0.8436 / \textbf{0.097} 
&  \underline{37.60} /  \underline{0.9616} 
& 28.04 / 0.8037 \\ \bottomrule
\end{tabular}
  }

%% file: tables/2.tex
  \resizebox{\linewidth}{!}{

\begin{tabular}{lccc}
\toprule
\multirow{2}{*}{Method}  & Full-Image         & Center-Image & Corner-Image \\          
\cmidrule(lr){2-4}
& PSNR\(\uparrow\)/SSIM\(\uparrow\)/LPIPS\(\downarrow\)    & PSNR/SSIM    & PSNR/SSIM    \\ \midrule

RCAN-\(\ell\)                                                                                      & 26.44 / 0.8676 / 0.147 & 26.91 / 0.8704 & 26.33 / 0.8667 \\

SwinIR-\(\ell\)                                                                                     & 26.14 / 0.8601 / 0.157 & 26.35 / 0.8612 & 26.11 / 0.8597 \\

ESRGAN                                                                                    & 25.78 / 0.8622 / 0.152 & 25.91 / 0.8637 & 25.77 / 0.8617 \\

BSRGAN                                                                                     & 24.77 / 0.8227 / 0.202 & 25.01 / 0.8233 & 24.71 / 0.8225 \\ \midrule

TTSR-\(\ell\)                                                                                       & 26.48 / 0.8676 / 0.147 & 27.69 / 0.8810 & 26.17 / 0.8631 \\

MASA-\(\ell\)                                                                                       & 26.36 / 0.8592 / 0.160 & 26.85 / 0.8620 & 26.25 / 0.8582 \\

DATSR-\(\ell\)                                                                                      & 26.17 / 0.8583 / 0.157 & 26.48 / 0.8596 & 26.11 / 0.8579 \\ \midrule

DCSR-\(\ell\)                                                                                        & 26.77 / 0.8748 / 0.134   & 28.87 / 0.9078 & 26.29 / 0.8635 \\
DCSR                                                                                    & 26.19 / 0.8553 / 0.110   & 28.05 / 0.8929 & 25.75 / 0.8425 \\

SelfDZSR-\(\ell\)                                                                                   & 26.27 / 0.8559 / 0.158 & 26.97 / 0.8591 & 26.10 / 0.8548 \\
SelfDZSR                                                                              & 25.98 / 0.8455 / 0.105 & 26.61 / 0.8496 & 25.81 / 0.8442 \\
ZeDuSR-\(\ell\)                                                                               & 25.41 / 0.8247 / 0.191 & 26.29 / 0.8336 & 25.21 / 0.8216 \\ 

KeDuSR-\(\ell\)                                                                                & \underline{27.66} / \underline{0.8890} / 0.117 & \underline{29.58} / 0.9303 & \underline{27.24} / \textbf{0.8750} \\
KeDuSR                                                                              & 27.18 / 0.8752 / \underline{0.084} & 29.06 / 0.9219 & 26.77 / 0.8593 \\
\midrule
Ours-\(\ell\)               
                                     
& \textbf{27.93} / \textbf{0.8900} /  0.117 
& \textbf{29.85} / \textbf{0.9358} 
& \textbf{27.50} / \underline{0.8743} \\

Ours           
                                                
& 27.36 / 0.8739 / \textbf{0.077} 
& 29.51 / \underline{0.9318} 
& 26.89 / 0.8542 
\\ \bottomrule
\end{tabular}
  }

%% file: tables/5.tex
\resizebox{1\linewidth}{!}{
\begin{tabular}{lcccccc}
\toprule
Method &
NRQM\(\uparrow\) &
NIQE\(\downarrow\) &
PI\(\downarrow\) &
BRISQUE\(\downarrow\) &
MUSIQ\(\uparrow\) &
MANIQA\(\uparrow\) \\ \midrule

RCAN-\(\ell\)        & 4.929 & 5.590 & 5.391 & 43.115 & 53.149 & 0.3295\\
SwinIR-\(\ell\)      & 4.763 & 6.066 & 5.714 & 42.262 & 51.087 & 0.3154 \\
ESRGAN-\(\ell\)      & 4.655 & 6.102 & 5.775 & 47.831 & 52.164 & 0.3223\\
BSRGAN-\(\ell\)      & 4.763 & 5.865 & 5.601 & 46.540 & 53.624 & 0.3379\\ \midrule
TTSR-\(\ell\)        & 4.655 & 4.879 & 4.964 & 39.636 & 55.526 & 0.3507\\
MASA-\(\ell\)        & 4.925 & 5.202 & 5.397 & 44.265 & 53.118 & 0.3274\\
DATSR-\(\ell\)       & 4.899 & 5.809 & 5.501 & 43.722 & 53.202 & 0.3292\\\midrule
DCSR-\(\ell\)        & 4.984 & 4.946 & 5.187 & 42.815 & 54.441 & 0.3392\\
SelfDZSR-\(\ell\)    & 4.862 & 5.716 & 5.472 & 45.607 & 54.709 & 0.3449\\
ZeDuSR    & 5.084    & 7.873    &6.462    &45.140     & 43.584 & 0.2942\\ 
KeDuSR-\(\ell\)      & \underline{5.268} & \underline{4.490} & \underline{4.682} & \underline{39.457} & \underline{58.506}  & \underline{0.3501}  \\ \midrule
Ours-\(\ell\)      & \textbf{5.434} & \textbf{4.127} & \textbf{4.432} & \textbf{37.020} & \textbf{59.236} & \textbf{0.3547}
\\ \bottomrule
\end{tabular}
}

%% file: tables/12.tex
\resizebox{\linewidth}{!}{
\begin{tabular}{
l@{\hspace{4pt}}
c@{\hspace{10pt}}
cccc
@{\hspace{8pt}}c
@{\hspace{2pt}}c
}
\toprule
\multirow{2}{*}{Method}
& \multirow{2}{*}{\makebox[0pt][c]{Adapt.}}
& \multicolumn{2}{c}{Paired}
& \multicolumn{2}{c}{Unpaired}
& \multirow{2}{*}{Params.}
& \multirow{2}{*}{Time} \\
\cmidrule(lr){3-4}
\cmidrule(lr){5-6}
&
& PSNR$\uparrow$
& SSIM$\uparrow$
& PSNR$\uparrow$
& SSIM$\uparrow$
& (M)
& (ms) \\
\midrule

\multirow{2}{*}{SPyNet}
& --
& 25.51
& 0.8766
& 20.60
& 0.7642
& \multirow{2}{*}{1.4}
& \multirow{2}{*}{20} \\

& $\checkmark$
& 33.78 {\scriptsize $(+8.3)$}
& 0.9470 {\scriptsize $(+.070)$}
& 23.59 {\scriptsize $(+3.0)$}
& 0.8075 {\scriptsize $(+.043)$}
&
& \\

\midrule

\multirow{2}{*}{PWC-Net}
& --
& 28.60
& 0.9289
& 21.46
& 0.7972
& \multirow{2}{*}{9.4}
& \multirow{2}{*}{16} \\

& $\checkmark$
& 33.72 {\scriptsize $(+5.1)$}
& 0.9504 {\scriptsize $(+.022)$}
& 23.47 {\scriptsize $(+2.0)$}
& 0.8082 {\scriptsize $(+.011)$}
&
& \\

\midrule

\multirow{2}{*}{RAFT}
& --
& 27.76
& 0.8811
& 20.84
& 0.7440
& \multirow{2}{*}{5.3}
& \multirow{2}{*}{64} \\

& $\checkmark$
& 34.40 {\scriptsize $(+6.6)$}
& 0.9549 {\scriptsize $(+.074)$}
& 23.41 {\scriptsize $(+2.6)$}
& 0.8074 {\scriptsize $(+.063)$}
&
& \\

\midrule

\multirow{2}{*}{GMFlow}
& --
& 27.82
& 0.8745
& 20.59
& 0.7213
& \multirow{2}{*}{4.7}
& \multirow{2}{*}{5.2} \\

& $\checkmark$
& 28.23 {\scriptsize $(+0.4)$}
& 0.8566 {\scriptsize $(-.018)$}
& 19.31 {\scriptsize $(-1.28)$}
& 0.6019 {\scriptsize $(-.119)$}
&
& \\

\midrule

\multirow{2}{*}{SEA-RAFT}
& --
& 28.94
& 0.9249
& 21.66
& 0.7894
& \multirow{2}{*}{19.66}
& \multirow{2}{*}{58} \\

& $\checkmark$
& \underline{34.53} {\scriptsize $(+5.6)$}
& \underline{0.9565} {\scriptsize $(+.032)$}
& \underline{23.68} {\scriptsize $(+2.0)$}
& 0.8158 {\scriptsize $(+.026)$}
&
& \\

\midrule

\multirow{2}{*}{WAFT}
& --
& 29.09
& 0.9271
& 21.68
& 0.7954
& \multirow{2}{*}{136}
& \multirow{2}{*}{212} \\

& $\checkmark$
& \textbf{35.59} {\scriptsize $(+6.5)$}
& \textbf{0.9623} {\scriptsize $(+.035)$}
& \textbf{24.12} {\scriptsize $(+2.4)$}
& \textbf{0.8310} {\scriptsize $(+.036)$}
&
& \\

\midrule

DuFlowNet
& $\checkmark$
& 33.74
& 0.9510
& 23.29
& \underline{0.8164}
& \textbf{0.65}
& \textbf{5} \\

\bottomrule
\end{tabular}
}

%% file: tables/8.tex







\scriptsize
\setlength{\tabcolsep}{1.6pt}
\renewcommand{\arraystretch}{1.10}

\resizebox{\columnwidth}{!}{
\begin{tabular}{c l c c c c}
\toprule
\multicolumn{3}{c}{Variants} 
& Full-Image  
& Center-Image 
& Corner-Image \\
\cmidrule(lr){1-3} \cmidrule(lr){4-6}
No. & Ref. Warping & Flow Prior
& PSNR\(\uparrow\)/SSIM\(\uparrow\)/LPIPS\(\downarrow\)
& PSNR/SSIM
& PSNR/SSIM \\
\midrule

\#1
& Patch Warping
& w/o flow
& 29.07 / 0.8592 / 0.204
& 35.07 / 0.9330 
& 28.13 / 0.8343 \\

\#2
& Patch Warping 
& SPyNet
& 29.14 / 0.8652 / 0.195
& 37.22 / 0.9575
& 28.09 / 0.8341 \\

\#3
& Patch Warping 
& DuFlowNet
& 29.22 / 0.8682 / 0.191
& 37.64 / 0.9637
& 28.15 / 0.8359 \\

\#4
& Patch + DCN
& DuFlowNet
& 29.60 / 0.8677 / 0.194        
& 37.40 / 0.9628
& 28.55 / 0.8356  \\

\#5
& Patch + KF-DCN
& DuFlowNet
& \textbf{29.76} / \textbf{0.8711} / 0.193                               
& \textbf{38.43} / \textbf{0.9683}  
& \textbf{28.68} / \textbf{0.8383}  \\

\bottomrule
\end{tabular}
}

%% file: tables/9.tex
  \resizebox{\linewidth}{!}{

\begin{tabular}{lccc}
\toprule
\multirow{2}{*}{\hspace{2em}Variants}  & Full-Image  & Center-Image & Corner-Image \\
\cmidrule(lr){2-4}
 &PSNR\(\uparrow\)/SSIM\(\uparrow\)/LPIPS\(\downarrow\)    
 & PSNR/SSIM    & PSNR/SSIM    \\ \midrule

\#1:Patch warping
& 29.22 / 0.8682 / 0.191
& 37.64 / 0.9637
& 28.15 / 0.8359 \\

\#2:KF-DCN warping
& 29.68 / 0.8686 / 0.197
& 37.21 / 0.9622
& 28.65 / 0.8370 \\

\#3:Region-concat
& 29.34 / 0.8673 / 0.195                               
& 37.24 / 0.9609  
& 28.30 / 0.8358  \\

\#4:Channel-concat
& 29.59 / 0.8684 / 0.195                               
& 38.13 / 0.9663  
& 28.52 / 0.8353  \\

\#5:Uniform aggregation
& 29.64 / 0.8692 / 0.195                             
& 37.42 / 0.9625
& 28.61 / 0.8376  \\

\#6:Ours
& \textbf{29.76} / \textbf{0.8711} / 0.193                               
& \textbf{38.43} / \textbf{0.9683}  
& \textbf{28.68} / \textbf{0.8383}  \\
\bottomrule
\end{tabular}
  }

%% file: supplementary.tex





\markboth{DuSRFlow: Supplementary Material}
{DuSRFlow: Supplementary Material}

\twocolumn[{
\begin{center}
{\Large\bfseries
DuSRFlow: Dual-Lens Super-Resolution via
Parallax-Aware Flow Estimation and
Complementary Reference Warping\par}

\vspace{0.8em}

{\large\bfseries Supplementary Material\par}

\vspace{1.5em}
\end{center}
}]

\setcounter{section}{0}
\setcounter{subsection}{0}
\setcounter{subsubsection}{0}

\renewcommand{\thesection}{S\arabic{section}}
\renewcommand{\thesubsection}{\thesection.\arabic{subsection}}
\renewcommand{\thesubsubsection}{\thesubsection.\arabic{subsubsection}}

\setcounter{equation}{0}
\renewcommand{\theequation}{S\arabic{equation}}

\setcounter{figure}{0}
\renewcommand{\thefigure}{S\arabic{figure}}

\setcounter{table}{0}
\renewcommand{\thetable}{S\arabic{table}}

\section{Implementation Details}
\subsection{Training Settings}
All experiments for DuFlowNet and DuSRFlow are implemented using the PyTorch framework and conducted on NVIDIA GeForce RTX 4090D GPUs. We employ the Adam optimizer with a cosine annealing strategy, where the initial learning rate is set to $10^{-4}$ and gradually decays to $10^{-6}$.
For the training of DuFlowNet, the batch size is set to 4, and the input LR patches are cropped to a size of $448\times448$. The foreground patch sizes are uniformly sampled between 80 and 200. For DuSRFlow training, we set the batch size to 4 and the patch size of input LR images to $256\times256$. We first train the DuFlowNet. After convergence, we further train the DuSRFlow network and fine-tune DuFlowNet. 

\subsection{Loss Functions}

\subsubsection{DuFlowNet}
DuFlowNet is trained with a photometric reconstruction loss and an edge-aware smoothness regularization. The photometric term is defined by the Charbonnier loss:
\begin{equation}
\mathcal{L}_{\text{f}}=
\sqrt{\left\|\mathbf{I}^{\text{HRC}}_{\downarrow}-\mathbf{I}^{\text{Ref}}_{\text{W}}\right\|_2^2+\varepsilon},
\end{equation}
where $\mathbf{I}^{\text{Ref}}_{\text{W}}$ denotes the warped reference image and $\mathbf{I}^{\text{HRC}}_{\downarrow}$ is the downsampled HR-center.

To suppress flow variations while preserving motion boundaries, we use a second-order edge-aware smoothness loss:
\begin{equation}
\mathcal{L}_{\text{S}}=
\sum_{i,j}\sum_{d\in\{x,y\}}
\left|\partial_d^2 \mathbf{f}(i,j)\right|
\,e^{-\alpha\left|\partial_d \mathbf{I}^{\text{LRC}}(i,j)\right|},
\end{equation}
where $\partial_d^2$ denotes the second-order spatial derivative along direction $d\in\{x,y\}$, $\mathbf{f}(i,j)$ is the predicted flow vector at pixel $(i,j)$, and $\alpha=10$ controls the edge-weighting strength.

The overall training objective for DuFlowNet is
\begin{equation}
\mathcal{L}_{\text{flow}}=
\lambda_{\text{f}}\mathcal{L}_{\text{f}}+
\lambda_{\text{S}}\mathcal{L}_{\text{S}},
\end{equation}
where $\lambda_{\text{f}}=1$ and $\lambda_{\text{S}}=0.5$.

\subsubsection{DuSRFlow}
The final DuSRFlow model is optimized with a hybrid objective consisting of reconstruction, perceptual, and adversarial terms:
\begin{equation}
\mathcal{L}_{\text{DuSR}}=
\mathcal{L}_{\text{ch}}+
\lambda_1\mathcal{L}_{\text{per}}+
\lambda_2\mathcal{L}_{\text{adv}}.
\end{equation}
The reconstruction term is defined as the Charbonnier loss:
\begin{equation}
\mathcal{L}_{\text{ch}}=
\sqrt{\left\|\mathbf{I}^{\text{HR}}-\mathbf{I}^{\text{SR}}\right\|_2^2+\varepsilon}.
\end{equation}
The perceptual loss is
\begin{equation}
\mathcal{L}_{\text{per}}=
\left\|\phi_i(\mathbf{I}^{\text{HR}})-\phi_i(\mathbf{I}^{\text{SR}})\right\|_2,
\end{equation}
where $\phi_i(\cdot)$ denotes the feature map extracted from the $i$-th layer of a pretrained VGG19 network. The adversarial loss $\mathcal{L}_{\text{adv}}$ is used to enhance visual realism. In all experiments, we set $\lambda_1=10^{-3}$ and $\lambda_2=10^{-4}$, and evaluate both reconstruction-only and hybrid training settings.

\section{Additional Experimental Results}

\subsection{Additional Quantitative Results}
We further present additional quantitative comparison results on RealMCVSR-Real and CameraFusion-Real.

As shown in Tables~\ref{tab:RealMCVSR-Real} and~\ref{tab:CameraFusion-Real}, our approach achieves strong performance on both datasets, further demonstrating the effectiveness of the proposed method.
\begin{table}[h]
  \centering
  \caption{Quantitative comparisons on RealMCVSR-Real.}
    \input{tables/3}
  \label{tab:RealMCVSR-Real}
\end{table}

\begin{table}[h]
  \centering
  \caption{ Quantitative comparisons on CameraFusion-Real.}
    \input{tables/4}
  \label{tab:CameraFusion-Real}
\end{table}

\subsection{Additional  Visual Results}

\begin{figure*}[!t]
  \centering
  \includegraphics[width=0.95\linewidth]{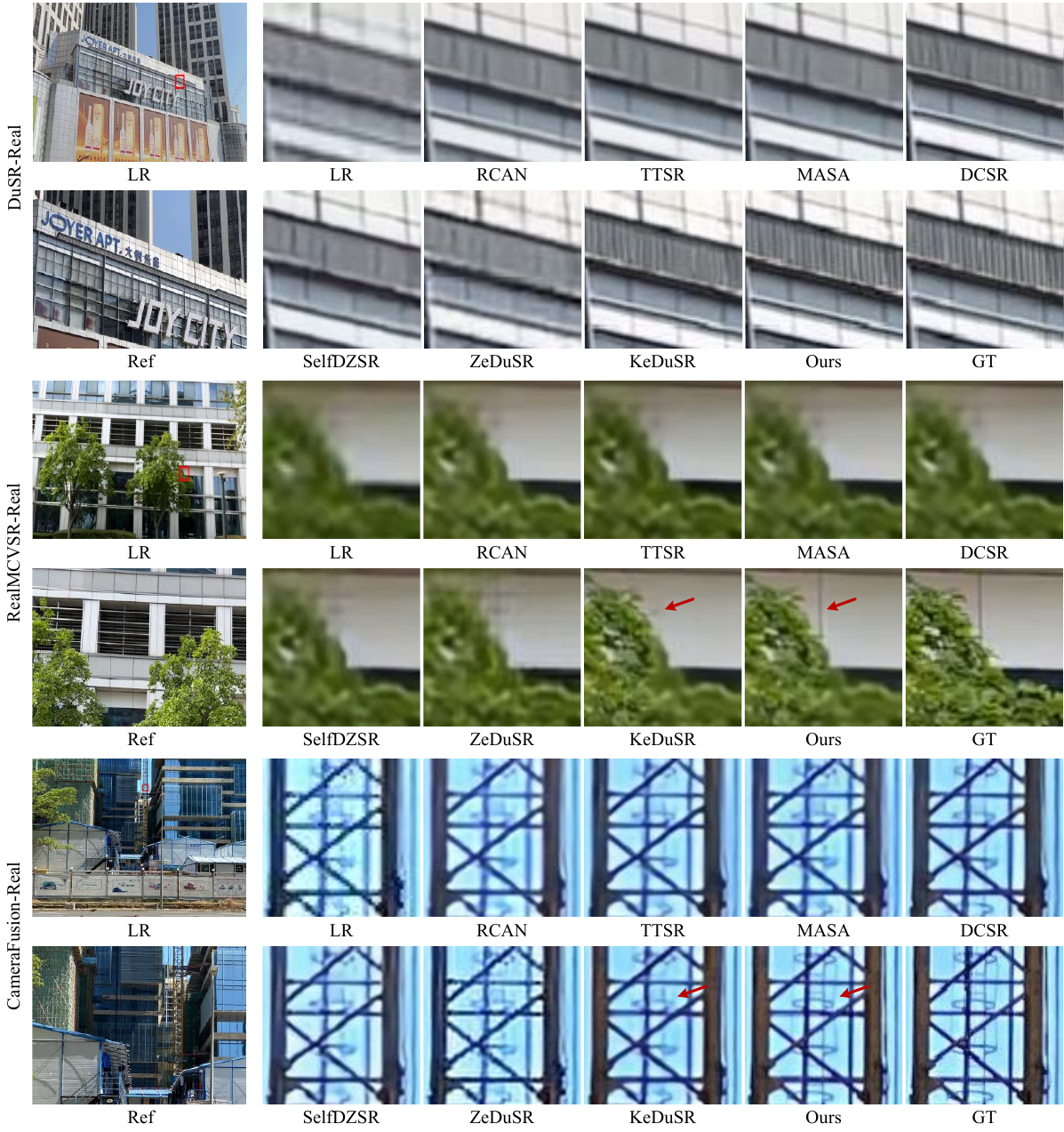}
   \caption{Visual comparisons on RealMCVSR-Real, DuSR-Real, and CameraFusion-Real datasets. Zoom in for better observation.}
   \label{fig:visual_additional}
\end{figure*}

\begin{figure*}
  \centering
  \includegraphics[width=0.85\linewidth]{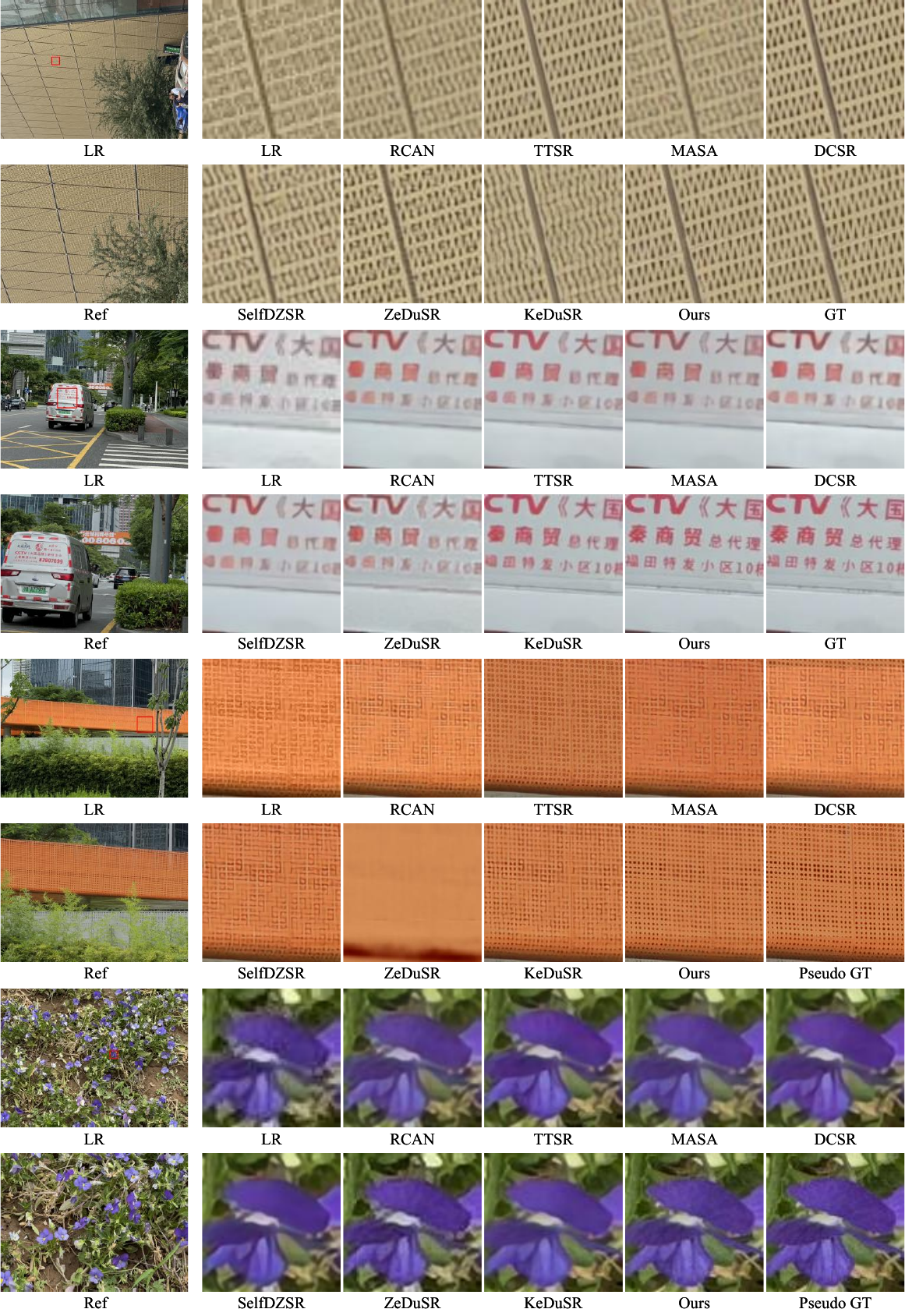}
   \caption{Visual comparisons of SR results on the DuSR-RealV2 dataset. The upper part shows results on the paired subset, and the lower part shows results on the unpaired subset. The presented results are obtained with only the reconstruction loss. For the unpaired subset, the pseudo ground truth is provided for reference only. Zoom in for better observation.}
   \label{fig:visual_dusrrealv2}
\end{figure*}

We further present additional visual comparison results on RealMCVSR-Real, DuSR-Real, and CameraFusion-Real datasets. As shown in Fig.~\ref{fig:visual_additional}, our approach reconstructs sharper structures and more faithful details across different real-world dual-lens datasets. In addition, Fig.~\ref{fig:visual_dusrrealv2} provides more visual comparisons on both the DuSR-RealV2-Paired and DuSR-RealV2-Unpaired subsets, further demonstrating the robustness of our method under different parallax and alignment conditions.

\subsection{Model Complexity}
We compare model complexity in Table~\ref{tab:model_complex}, including the number of parameters, FLOPs, and inference latency. FLOPs and latency are measured when generating an HR output of size $1792\times896$. For fair comparison with previous studies, all timing results are measured on a single NVIDIA RTX 3090 GPU. Owing to the lightweight design of DuFlowNet, DuSRFlow reduces both parameter count and inference latency compared with KeDuSR, while achieving the second-fastest inference speed among all compared RefSR methods.

\subsection{Cross-Dataset Generalization}
To evaluate the generalization ability of methods trained on the DuSR-RealV2 dataset, we test them on additional benchmark datasets. As shown in Table~\ref{tab:generalization}, our method achieves the best generalization performance.
\begin{table}[h]
  \centering
\caption{Generalization evaluation of models trained on DuSR-RealV2.}
\input{tables/10}
  \label{tab:generalization}
\end{table}

\section{Ablation on DuFlowNet}

\rev{To validate the proposed real--synthetic data fusion strategy, we compare DuFlowNet trained without parallax simulation, solely on synthetic pairs, and with the complete strategy. The \textit{w/o parallax sim.} variant removes the foreground-based parallax simulation, while the \textit{Synthetic-only} variant replaces the real dual-lens backgrounds with DIV2K images using bicubic downsampling and random homography transformations. Alignment consistency is evaluated using PSNR and SSIM against the HR-center image on the Paired subset and the LR-center image on the Unpaired subset.}

\begin{table}[H]
  \centering

\caption{Ablation study of our real–synthetic data fusion strategy.}
\input{tables/sup1}
  \label{tab:Ablation Flownet}
\end{table}

\begin{figure}[H]
\centering
\includegraphics[width=0.75\linewidth]{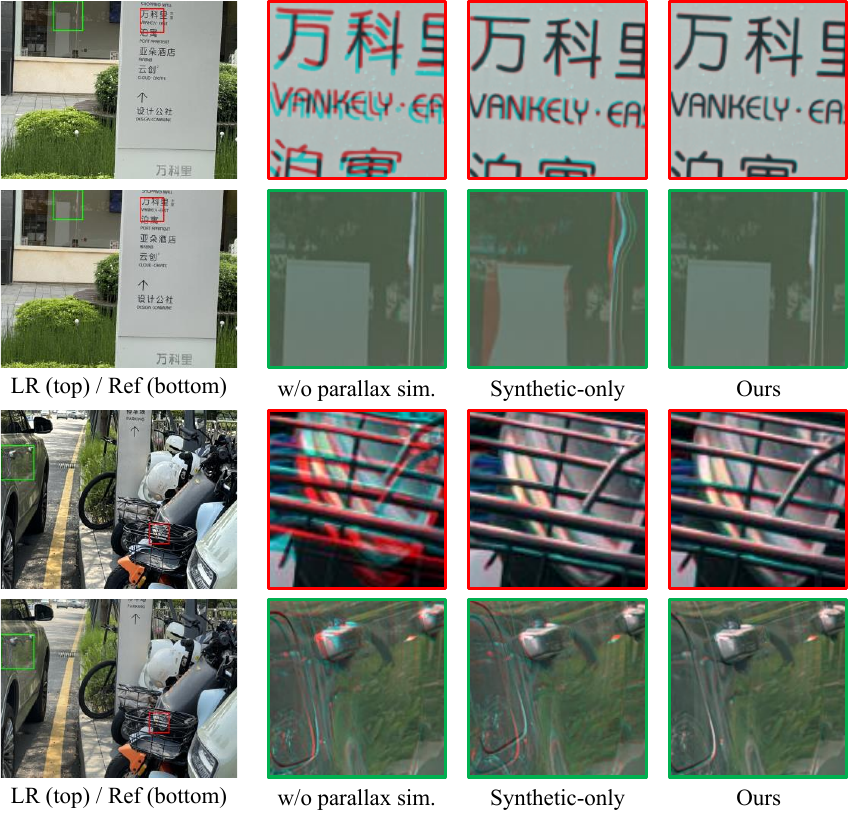}
\caption{\rev{Visual comparison of alignment results. The red and green boxes highlight large-parallax foreground and background regions, respectively.}}
   \label{sec:ablation flownet}
\end{figure}

\rev{As reported in Table~\ref{tab:Ablation Flownet}, the \textit{w/o parallax sim.} variant performs poorly on the Unpaired subset, since foreground simulation provides controllable large-parallax samples that are particularly important for this challenging subset. In contrast, the \textit{Synthetic-only} variant performs worst on the Paired subset because its synthesized backgrounds cannot fully reproduce the characteristics of real dual-lens images. The visual results in Fig.~\ref{sec:ablation flownet} further support these observations. In the red-boxed large-parallax regions, the model without foreground simulation exhibits evident misalignment, whereas in the green-boxed background regions, the \textit{Synthetic-only} model produces larger alignment errors. By combining real dual-lens backgrounds with controllable foreground parallax simulation, our complete strategy achieves more reliable alignment in both regions.}

\section{Memory Overhead}

\rev{We report the inference-time peak GPU memory for reconstructing a
$1792\times896$ SR image. The two warping branches are executed
sequentially and share the KF indices and confidence map. As shown in
Table~\ref{tab:memory_overhead}, the complete model requires 3600.75 MiB,
only 98 MiB ($2.8\%$) more than KF-DCN warping alone, keeping the peak
memory close to the more demanding branch rather than the sum of both
branches.}

\begin{table}[H]
\centering

\caption{\rev{Inference-time peak GPU memory of different warping
configurations.}}
\label{tab:memory_overhead}
{
\resizebox{0.7\linewidth}{!}{
\begin{tabular}{lc}
\toprule
Variant & Peak Memory (MiB) \\
\midrule
Patch warping only & 2371.09 \\
KF-DCN warping only & 3502.75 \\
Ours & 3600.75 \\
\bottomrule
\end{tabular}
}}
\end{table}

\begin{table*}[t]
  \centering
\caption{FLOPs and latency are measured for generating an HR output of size $1792\times896$ on a single NVIDIA RTX 3090 GPU.}
\input{tables/6}
  \label{tab:model_complex}
\end{table*}

%% file: tables/3.tex
  \resizebox{\linewidth}{!}{
\begin{tabular}{lccc}
\toprule
\multirow{2}{*}{Method} & Full-Image         & Center-Image & Corner-Image \\
\cmidrule(lr){2-4}
                        & PSNR\(\uparrow\)/SSIM\(\uparrow\)/LPIPS\(\downarrow\)    & PSNR/SSIM    & PSNR/SSIM    \\ \midrule
 
RCAN-\(\ell\)                    & 25.96 / 0.8033 / 0.234 & 25.69 / 0.7937 & 26.12 / 0.8065 \\
 
SwinIR-\(\ell\)                  & 25.78 / 0.7982 / 0.246 & 25.50 / 0.7885 & 25.94 / 0.8015 \\ \midrule
 
TTSR-\(\ell\)                    & 25.92 / 0.8017 / 0.235 & 25.94 / 0.7962 & 25.98 / 0.8036 \\
 
MASA-\(\ell\)                    & 25.95 / 0.7989 / 0.239 & 25.81 / 0.7899 & 26.07 / 0.8020 \\
 
DATSR-\(\ell\)                   & 25.81 / 0.7975 / 0.242 & 25.58 / 0.7882 & 25.95 / 0.8007 \\ \midrule
 
DCSR-\(\ell\)                    & 26.28 / 0.8111 / 0.217 & 27.19 / 0.8298 & 26.08 / 0.8048 \\
DCSR                & 25.85 / 0.7966 / 0.186 & 26.98 / 0.8476 & 25.58 / 0.7793 \\
 
SelfDZSR-\(\ell\)                & 25.33 / 0.7928 / 0.246 & 25.66 / 0.7860 & 25.30 / 0.7952 \\
SelfDZSR            & 25.24 / 0.7786 / 0.175 & 25.50 / 0.7732 & 25.23 / 0.7805 \\ 
ZeDuSR-\(\ell\)        & 24.98 / 0.7702 / 0.262 & 25.38 / 0.7650 & 24.93 / 0.7720 \\ 
 
KeDuSR-\(\ell\)              
& \underline{27.05} / \underline{0.8406} / 0.180 
& \underline{29.25} / 0.9191
& \underline{26.56} / \textbf{0.8139} \\

KeDuSR           
& 26.42 / 0.8184 / \underline{0.127}
& 28.51 / 0.9090 
& 25.95 / 0.7875 \\ \midrule

Ours-\(\ell\)               
& \textbf{27.16} / \textbf{0.8416} /  0.181
& \textbf{29.54} / \textbf{0.9279} 
& \textbf{26.64} / \underline{0.8122} \\

Ours          
& 26.46 / 0.8182 / \textbf{0.126} 
& 29.04 / \underline{0.9221}
& 25.90 / 0.7828

\\ \bottomrule
\end{tabular}
  }

%% file: tables/4.tex
  \resizebox{\linewidth}{!}{
\begin{tabular}{lccc}
\toprule
                         & Full-Image         & Center-Image & Corner-Image \\
                         \cmidrule(lr){2-4}
\multirow{-2}{*}{Method} & PSNR\(\uparrow\)/SSIM\(\uparrow\)/LPIPS\(\downarrow\)    & PSNR/SSIM    & PSNR/SSIM    \\ \midrule
 
RCAN-\(\ell\)                     & 25.67 / 0.8049 / 0.308 & 26.65 / 0.8158 & 25.45 / 0.8012 \\

SwinIR-\(\ell\)                   & 25.32 / 0.8007 / 0.315 & 25.81 / 0.8073 & 25.22 / 0.7985 \\ \midrule

TTSR-\(\ell\)                     & 25.83 / 0.8044 / 0.311 & 26.75 / 0.8188 & 25.62 / 0.7996 \\
 
MASA-\(\ell\)                     & 25.78 / 0.8030 / 0.303 & 26.70 / 0.8155 & 25.58 / 0.7988 \\ \midrule

DCSR-\(\ell\)                     & 26.02 / 0.8123 / 0.293   & 28.37 / 0.8440 & 25.51 / 0.8016 \\
DCSR                 & 25.47 / 0.7605 / 0.165   & 27.14 / 0.7883 & 25.08 / 0.7512 \\
SelfDZSR-\(\ell\)                 & 25.94 / 0.8041 / 0.283 & 27.10 / 0.8148 & 25.68 / 0.8005 \\
SelfDZSR             & 25.64 / 0.7790 / 0.151 & 26.77 / 0.7897 & 25.39 / 0.7753 \\ 
ZeDuSR-\(\ell\)        & 26.16 / 0.7920 / 0.279   & 27.44 / 0.8067 & 25.87 / 0.7871 \\ 

KeDuSR-\(\ell\)                
& \underline{27.53} / \underline{0.8292} / 0.276 
& 30.48 / 0.8656 
& \underline{26.93} / \textbf{0.8169} 
\\

KeDuSR            
& 27.00 / 0.7931 / \underline{0.133} 
& 29.77 / 0.8418 
& 26.43 / 0.7768 
\\  \midrule

Ours-\(\ell\)                
& \textbf{27.71} / \textbf{0.8391} /  0.258
& \textbf{31.84} / \textbf{0.9055} 
& \textbf{26.96} / \underline{0.8168} \\

Ours
& 27.24 /  0.8105 / \textbf{0.122} 
& \underline{31.58} /  \underline{0.9016}
& 26.45 /  0.7799\\ \bottomrule

\end{tabular}
  }

%% file: tables/10.tex
\resizebox{\linewidth}{!}{
\begin{tabular}{lccc}
\toprule
\multirow{2}{*}{Method} 
& DuSR-Real 
& RealMCVSR-Real 
& CameraFusion-Real \\ 
\cmidrule(lr){2-4}
& PSNR\(\uparrow\)/SSIM\(\uparrow\)/LPIPS\(\downarrow\) 
& PSNR\(\uparrow\)/SSIM\(\uparrow\)/LPIPS\(\downarrow\) 
& PSNR\(\uparrow\)/SSIM\(\uparrow\)/LPIPS\(\downarrow\) \\ 
\midrule

TTSR-\(\ell\)        
& 24.59 / 0.8374 / 0.175 
& 23.15 / 0.7513 / 0.273 
& 25.58 / 0.8018 / 0.279 \\

MASA-\(\ell\)        
& 24.84 / 0.8270 / 0.198 
& 24.40 / 0.7575 / 0.285 
& 25.43 / 0.7947 / 0.289 \\

DATSR-\(\ell\)        
& 24.58 / 0.8278 / 0.196 
& 24.01 / 0.7610 / 0.271 
& 25.17 / 0.7917 / 0.288 \\

DCSR-\(\ell\)        
& 25.30 / 0.8497 / 0.169
& 24.69 / 0.7807 / 0.251 
& 25.56 / 0.8054 / 0.276 \\

SelfDZSR-\(\ell\)    
& 25.79 / 0.8402 / 0.188 
& 24.82 / 0.7703 / 0.274 
& 26.35 / 0.8057 / 0.290 \\

ZeDuSR-\(\ell\)      
& 26.66 / 0.7911 / 0.276 
& 24.98 / 0.7702 / 0.262 
& 26.16 / 0.7920 / 0.279 \\

KeDuSR-\(\ell\)      
& \underline{26.75} / \underline{0.8639} / \underline{0.149} 
& \textbf{25.94} / \underline{0.8117} / \underline{0.207} 
& \underline{27.37} / \underline{0.8232} / \underline{0.253} \\
\midrule

Ours-\(\ell\)      
& \textbf{26.94} / \textbf{0.8725} / \textbf{0.136} 
& \underline{25.89} / \textbf{0.8258} / \textbf{0.190} 
& \textbf{27.43} / \textbf{0.8354} / \textbf{0.218} \\

\bottomrule
\end{tabular}
}

%% file: tables/sup1.tex
\resizebox{0.8\linewidth}{!}{
\begin{tabular}{ccccc}
\toprule
Dataset & Metrics & w/o parallax sim. & Synthetic-only & Ours
\\ \midrule

\multirow{2}{*}{Paired}
& PSNR\(\uparrow\) & 33.72 & 31.15 & 33.74
\\
& SSIM\(\uparrow\) & 0.9507 & 0.9050 & 0.9510
\\

\midrule

\multirow{2}{*}{Unpaired}
& PSNR\(\uparrow\) & 22.43 & 22.83 & 23.29
\\
& SSIM\(\uparrow\) & 0.7788 & 0.7526 & 0.8164
\\

\bottomrule
\end{tabular}
}

%% file: tables/6.tex

\resizebox{0.90\textwidth}{!}{
\begin{tabular}{lcccccccccccc}
\toprule
 \multirow{2}{*}{\vspace{-1em}\hspace{1em}Method} & \multicolumn{4}{c}{SISR} & \multicolumn{8}{c}{RefSR} \\ 
\cmidrule(lr){2-5} \cmidrule(lr){6-13}
 & RCAN & SwinIR & ESRGAN & BSRGAN & TTSR & MASA & DATSR & DCSR & SelfDZSR & ZeDuSR & KeDuSR & Ours \\ \midrule
Latency (s) & 0.69 & 2.85 & 0.08 & 0.45 & 7.51 & 1.52 & 9.35 & 0.84 & 0.17 & 180 & 0.51 & 0.47 \\
Params (M) & 5.63 & 11.75 & 16.70 & 16.61 & 6.25 & 4.02 & 10.51 & 3.19 & 0.52 & 1.49 & 5.63 & 5.02 \\
FLOPs (T) & 6.15 & 0.41 & 1.80 & 6.77 & 2.92 & 2.34 & 5.81 & 2.08 & 1.14 & —— & 2.46 & 2.76 \\ \bottomrule
\end{tabular}
}